\documentclass{article}
\usepackage{ijcai25}

\usepackage{times}
\usepackage{soul}
\usepackage{url}
\usepackage[hidelinks]{hyperref}
\usepackage[utf8]{inputenc}
\usepackage[small]{caption}
\usepackage{graphicx}
\usepackage{amsmath}
\usepackage{amsthm}
\usepackage{booktabs}
\usepackage{algorithm}
\usepackage{algorithmic}
\usepackage[switch]{lineno}

\usepackage{subcaption}
\usepackage{alphalph}

\usepackage{amsfonts}
\usepackage{array}
\usepackage{textcomp}

\usepackage{verbatim}
\usepackage{pgfplots}
\pgfplotsset{compat=1.18}
\usepackage{multirow}
\usepackage{svg}
\usepackage{subcaption}
\usepackage{enumitem}
\usepackage{morefloats}
\usepackage{makecell}
\usepackage{color}

\newcommand{\revision}[1]{#1}

\newcommand{\citep}[1]{\citeauthor{#1}~\shortcite{#1}}

\title{Bridging Local and Global Knowledge via Transformer in Board Games\footnote{The full version is at \url{https://arxiv.org/abs/2410.05347}.}}

\author{
Yan-Ru Ju
\and
Tai-Lin Wu\and
Chung-Chin Shih\and
Ti-Rong Wu\thanks{Corresponding author.}\\
\affiliations
Institute of Information Science, Academia Sinica, Taiwan\\
\emails
\{yanru, qwertgb1234, rockmanray, tirongwu\}@iis.sinica.edu.tw
}

\begin{document}

\maketitle

\begin{abstract}
Although AlphaZero has achieved superhuman performance in board games, recent studies reveal its limitations in handling scenarios requiring a comprehensive understanding of the entire board, such as recognizing long-sequence patterns in Go.
To address this challenge, we propose \textit{ResTNet}, a network that interleaves residual and Transformer blocks to bridge local and global knowledge.
ResTNet improves playing strength across multiple board games, increasing win rate from 54.6\% to 60.8\% in 9x9 Go, 53.6\% to 60.9\% in 19x19 Go, and 50.4\% to 58.0\% in 19x19 Hex.
In addition, ResTNet effectively processes global information and tackles two long-sequence patterns in 19x19 Go, including \textit{circular pattern} and \textit{ladder pattern}.
It reduces the mean square error for circular pattern recognition from 2.58 to 1.07 and lowers the attack probability against an adversary program from 70.44\% to 23.91\%.
ResTNet also improves ladder pattern recognition accuracy from 59.15\% to 80.01\%.
By visualizing attention maps, we demonstrate that ResTNet captures critical game concepts in both Go and Hex, offering insights into AlphaZero's decision-making process.
Overall, ResTNet shows a promising approach to integrating local and global knowledge, paving the way for more effective AlphaZero-based algorithms in board games.
\revision{Our code is available at https://rlg.iis.sinica.edu.tw/papers/restnet.}
\end{abstract}




\section{Introduction}
AlphaZero \cite{silver_general_2018} has demonstrated superhuman performance in various board games, such as Go, chess, and Shogi.
Following the advent of AlphaZero, researchers have successfully reproduced and extended the algorithm, achieving superhuman performance across a broader range of board games, such as Hex \cite{gao_threehead_2018} and Gomoku \cite{xie_alphagomoku_2018,liang_alphazero_2023}.
However, despite these advancements, several studies \cite{wang_adversarial_2023,tian_elf_2019} have reported that AlphaZero remains vulnerable in scenarios requiring a more comprehensive understanding of the entire board, particularly in games with larger board sizes, such as 19x19 Go.

\begin{figure}[t]
\centering
\subfloat[Circular pattern]{
  \includegraphics[width=0.2\textwidth]{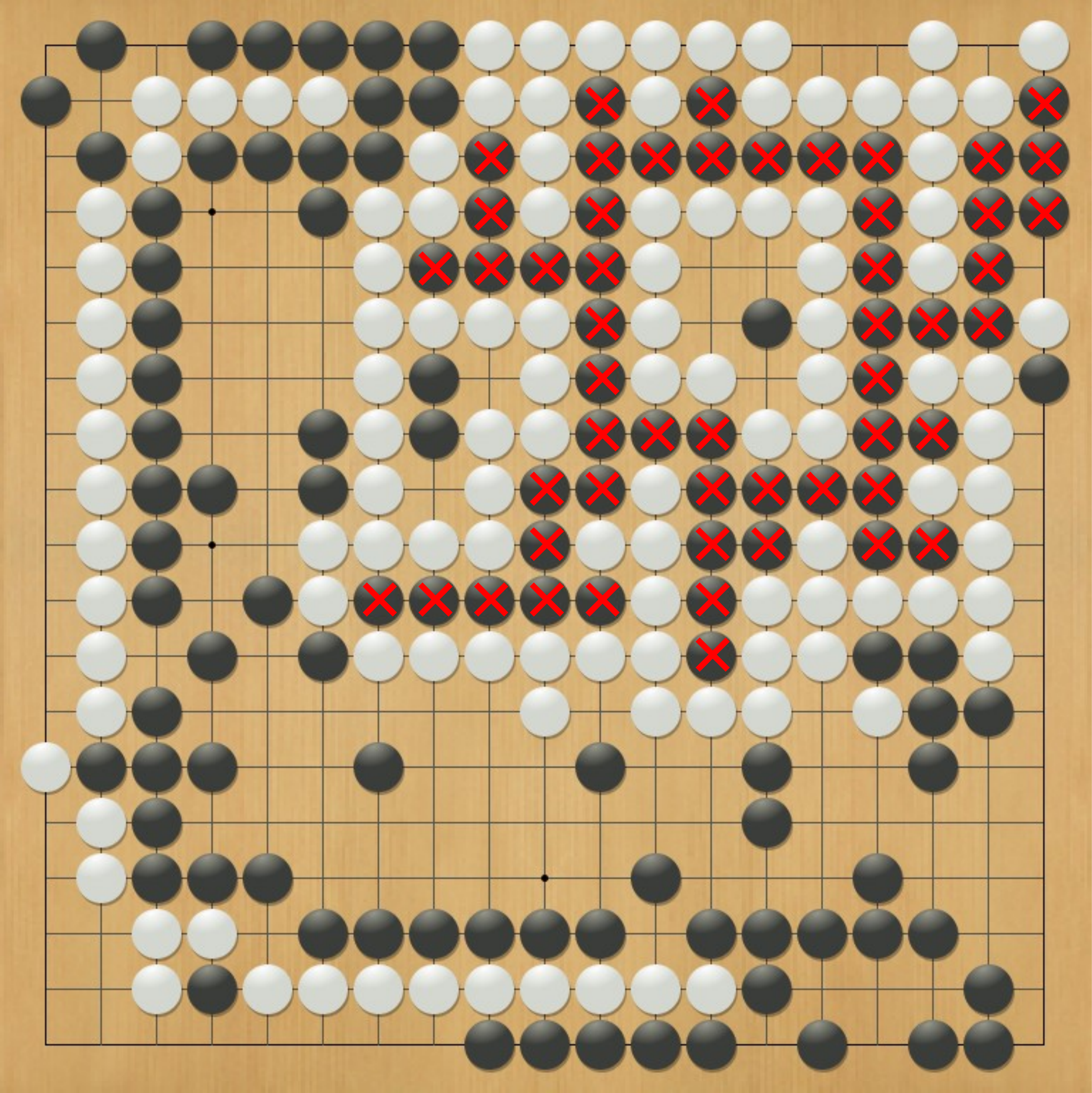}
  \label{fig:intro_circular}
}
\subfloat[Ladder pattern]{
  \includegraphics[width=0.2\textwidth]{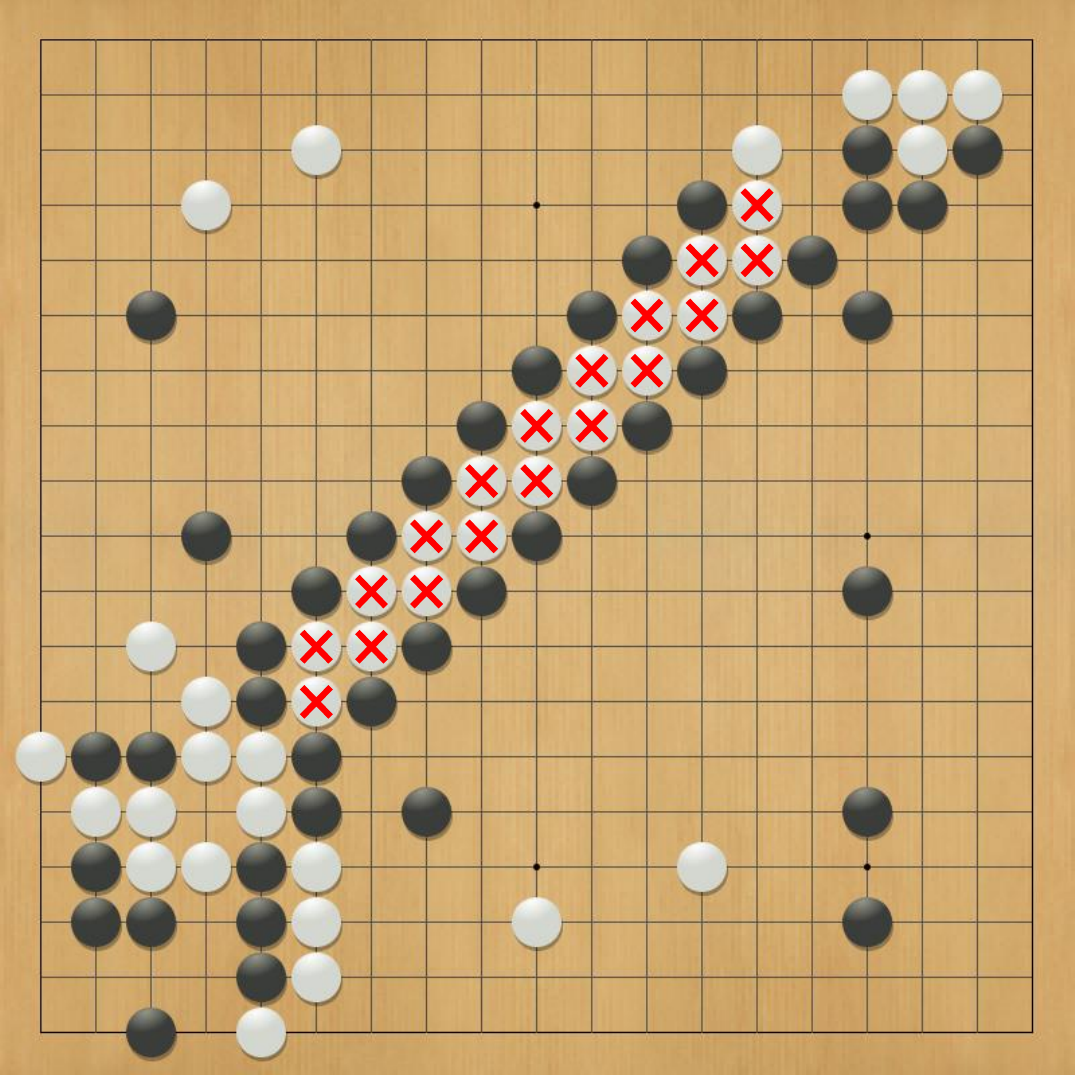}
  \label{fig:intro_ladder}
}
\caption{Two challenging long-sequence patterns that require an understanding of global knowledge in 19x19 Go.}
\label{fig:intro_ladder_circular}
\end{figure}

For example, recent research \cite{wang_adversarial_2023} introduces a specialized program, named \textit{cyclic-adversary}, designed specifically to find the weaknesses in KataGo \cite{wu_accelerating_2020a}, the current strongest open-sourced Go program.
During the game, the cyclic-adversary subtly induces KataGo to form a circular pattern, as illustrated by the marked black stones in Figure \ref{fig:intro_circular}, while simultaneously encircling and capturing these stones.
These circular patterns usually cover a wide range of the Go board, requiring a global understanding to handle these patterns effectively.
Consequently, the cyclic-adversary successfully attacks KataGo with a success rate exceeding 90\%.
Another challenge requiring global understanding is the recognition of the \textit{ladder pattern} in Go.
\revision{In ladder}, the player needs to simulate a long sequence of moves in zig-zag patterns across the entire board to capture or escape a group of stones, as shown by the marked white stones in Figure \ref{fig:intro_ladder}.
While this concept is relatively simple for human players, a previous study \cite{tian_elf_2019} has shown that AlphaZero struggles to handle ladder patterns perfectly.

These examples highlight that handling global information remains a significant challenge in AlphaZero algorithms for board games, primarily due to their reliance on convolutional neural networks (CNNs), which are inherently designed to capture local patterns.
To address this limitation, Transformers with self-attention mechanisms \cite{sutskever_sequence_2014,vaswani_attention_2017} present a promising alternative, offering strong capabilities in capturing the global context.
For example, similar to board games, computer vision -- particularly image classification -- primarily relies on extracting features in local patterns using CNNs.
Recently, several network architectures, such as Vision Transformers (ViT) \cite{dosovitskiy_image_2020}, Convolutional vision Transformers (CvT) \cite{wu_cvt_2021}, and CoAtNet \cite{dai_coatnet_2021}, have been proposed to leverage Transformers to better capture global patterns in computer vision.
However, directly applying these Transformers from computer vision to games does not yield better results \cite{czech_representation_2023,tseng_can_2024} due to the distinct objectives of the two domains: while the former focuses on achieving higher accuracy in image classification, the latter aims to maximize win rates in gameplay.
This raises an important research question: \textit{How can Transformers be effectively integrated into AlphaZero algorithms to improve global information processing while preserving the ability to recognize local patterns for board games?}

To answer this question, this paper introduces \textit{ResTNet}, a novel network architecture specifically designed for AlphaZero algorithms which interleaves the use of residual and Transformer blocks to effectively balance between local and global information processing.
The contribution of this paper can be summarized as follows:
\begin{itemize}[left=0pt]
    \item For the first time, the integration of Transformers and residual networks within AlphaZero for board games is thoroughly investigated.
    Experiments show ResTNet improves playing strength, with win rates increasing from 54.6\% to 60.8\% in 9x9 Go, 53.6\% to 60.9\% in 19x19 Go \revision{(using four KataGo models \cite{wu_accelerating_2020a} as baselines)}, and 50.4\% to 58.0\% in 19x19 Hex \revision{(using MoHex \cite{arneson_monte_2010} as the baseline)}, indicating a promising approach for training AlphaZero in board games.
    \item ResTNet demonstrates the ability to process global information by: (a) reducing the mean square error for recognizing circular patterns from 2.58 to 1.07 and lowering the attack probability from 70.44\% to 23.91\% against the cyclic-adversary among 24 games provided by \citep{wang_adversarial_2023}, and (b) improving ladder recognition accuracy from 59.15\% to 80.01\% in a human game collection.
    \item RestNet offers insights into the interpretation of AlphaZero models and enhances explainability by visualizing attention maps from Transformers, capturing key concepts in both Go and Hex.
\end{itemize}

\section{Background}
\label{sec:background}
\subsection{AlphaZero}
AlphaZero \cite{silver_general_2018} is a reinforcement learning-based algorithm that combines deep neural networks with search algorithms to master board games without the need for any human knowledge.
The network architecture is composed of several residual blocks with convolutional layers, followed by two head outputs: a policy head for predicting the probability distribution of the next move, and a value head for estimating the win rate. 
The training process consists of self-play and optimization phases.
In the self-play phase, AlphaZero executes a Monte Carlo Tree Search (MCTS) \cite{coulom_efficient_2007,browne_survey_2012} to generate self-play games and store them in a replay buffer.
In the optimization phase, self-play games are uniformly sampled from the replay buffer and used to optimize the neural network model.
The policy network aims to learn the move distribution that reflects the search results of the MCTS, while the value network aims to predict the game outcomes.
The newly optimized network models are then used to generate new self-play games.
Repeating this process allows AlphaZero to progressively enhance its performance and achieve superhuman performance.

\subsection{Transformers in Computer Vision}
\revision{In recent years, Transformers have achieved remarkable success in natural language processing \cite{vaswani_attention_2017,devlin_bert_2019}, and have also been successfully extended to computer vision.}
\revision{Unlike textual data, images contain rich positional and local information.
To accommodate this, the Vision Transformer (ViT) \cite{dosovitskiy_image_2020} divides an image into patches, treating each patch as a token analogous to a word in a sentence.}
This allows the network to apply the Transformer's sequential data processing capabilities to visual data.

Recent trends in computer vision show that integrating convolutional operations within the Transformer significantly enhances the performance \cite{wu_cvt_2021,wang_pyramid_2021,guo_cmt_2022,dai_coatnet_2021}.
\revision{For example, CoAtNet \cite{dai_coatnet_2021} proposes a network architecture that begins with several convolutional blocks, followed by a series of Transformer blocks with relative attention \cite{shaw_selfattention_2018}.}
CoAtNet has demonstrated state-of-the-art performance on image classification tasks, even under conditions of low data availability.

\subsection{Transformers in Board Games}
Transformers have recently been applied to board games, particularly in two directions: (a) using text representations of board games to frame game-playing AI as a natural language processing task, and (b) using visual representations of board games to apply Transformers for processing these representations, similar to image classification tasks.
For text representations, \citep{ciolino_go_2020} fine-tuned GPT-2 on Go games to predict the next move based on the SGF text format.
Similarly, \citep{feng_chessgpt_2024} proposed ChessGPT, which finetunes a pre-trained Large Language Model (LLM) on chess datasets, including games, books, and videos, to predict moves or evaluate positions.
\citep{ruoss_amortized_2024} trained a chess policy network through supervised learning on high-quality games annotated by Stockfish 16 \cite{romstad_stockfish_2023}, demonstrating that search-based programs can be distilled into large-scale transformers.
\citep{schultz_mastering_2024} introduced the Multi-Action-Value (MAV) model, which integrates Transformers with internal and external planning and supports multiple board games within a unified framework.

On the other hand, for visual representation, researchers utilize Transformers to process board states directly.
\citep{sagri_vision_2023} proposed replacing the residual network with EfficientFormer \cite{li_efficientformer_2022} and trained the policy and value network from KataGo's self-play games.
Their experiments show that using EfficientFormer can achieve a higher win rate in scenarios where only CPU is available.
Similarly, \citep{monroe_mastering_2024} applied ViT architecture to train policy and value networks by supervised learning from a self-play game collection in the chess environment, achieving higher win rates when using only the policy or value network.
In addition, \citep{czech_representation_2023} proposed combining Mobile Convolutional Blocks \cite{sandler_mobilenetv2_2018} and Next Transformer Blocks \cite{li_nextvit_2022} in the chess environment.
They improved the Transformer performance by identifying chess-specific input features.

While these studies have made progress in applying Transformers to visual board representations, they primarily adapt Transformers from computer vision without thoroughly investigating configurations tailored for board games.
Furthermore, most of these works focus on supervised learning on game collections rather than training within the reinforcement learning framework of the AlphaZero algorithm.
In contrast, our work focuses on visual representation by leveraging Transformers to design novel network configurations specifically for board games, bridging local and global knowledge within the AlphaZero training algorithm.

\section{ResTNet}
\label{sec:our_design}

\subsection{Network Design}
\label{network_design}
\begin{figure*}[ht!]
    \centering
    \subfloat[Network architecture of ResTNet]{
        \includegraphics[width=0.9\linewidth]{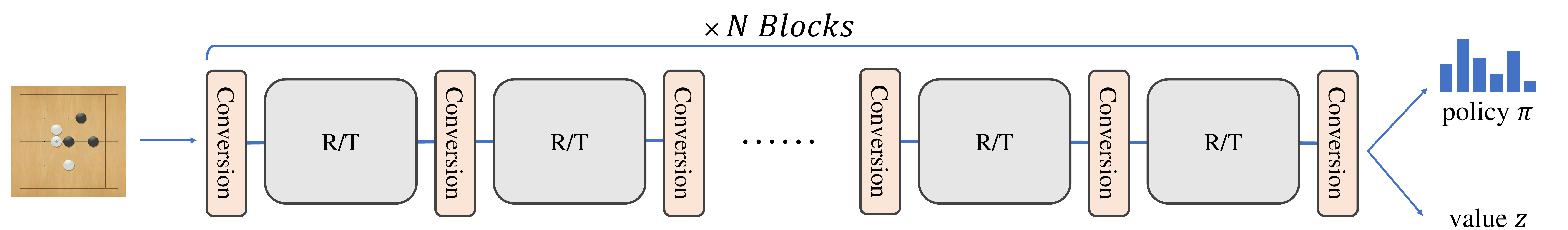}
        \label{fig:arch_restnet}
    }
    \\
    \subfloat[Residual block]{
        \includegraphics[width=0.20\linewidth]{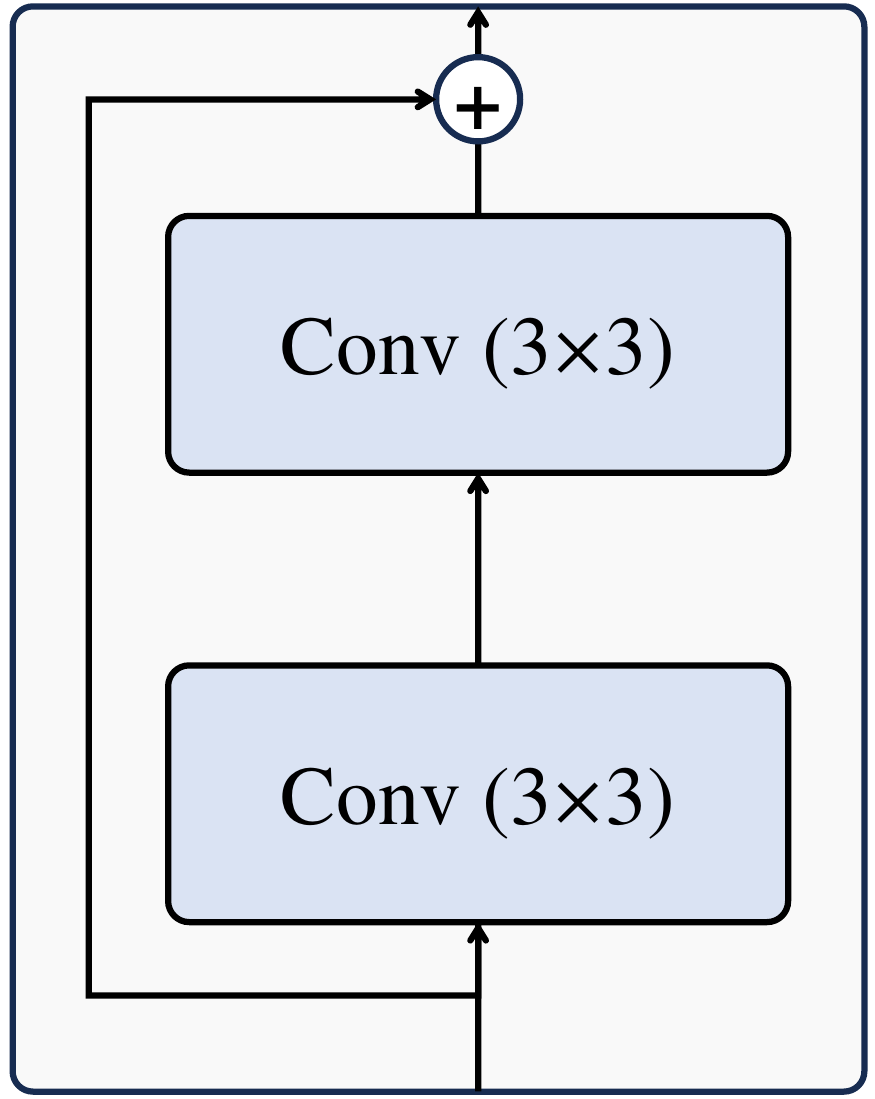}
        \label{fig:arch_r}
    }
    \subfloat[Transformer block]{
        \includegraphics[width=0.20\linewidth]{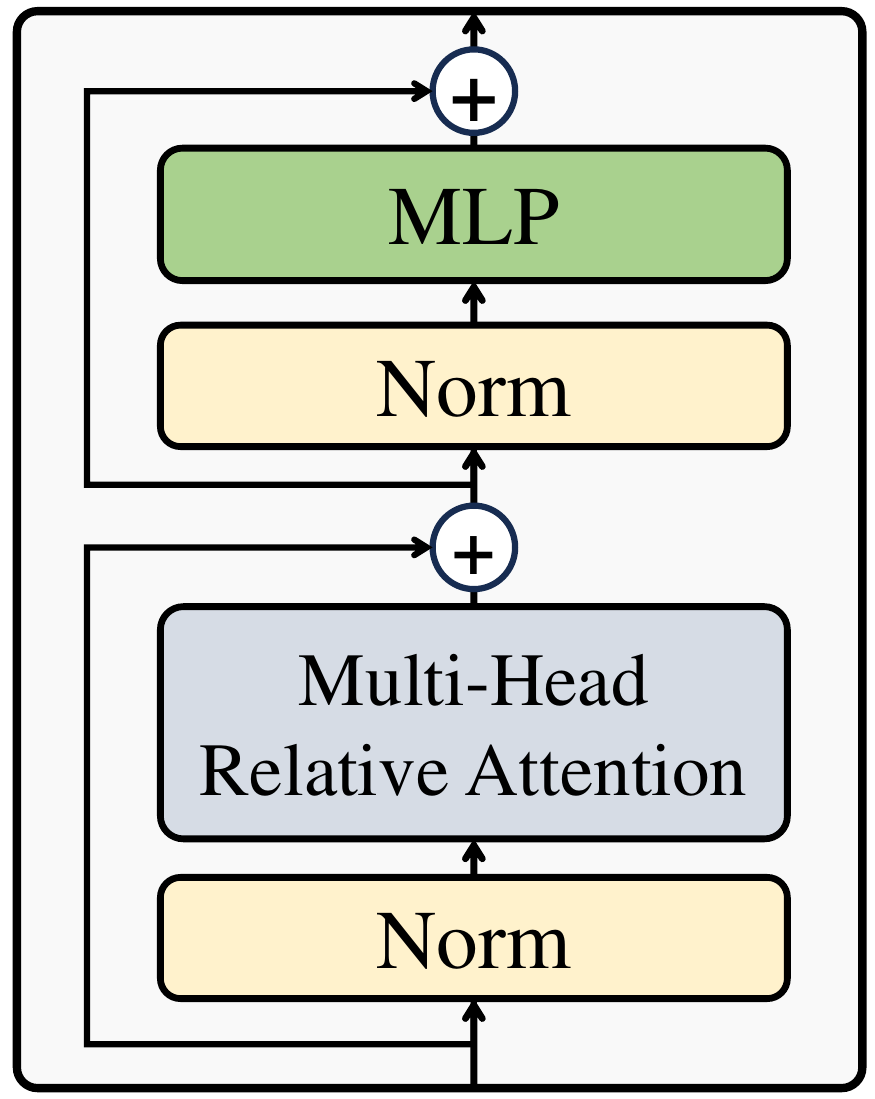}
        \label{fig:arch_t}
    }
    \subfloat[Feature conversion]{
        \includegraphics[width=0.53\linewidth]{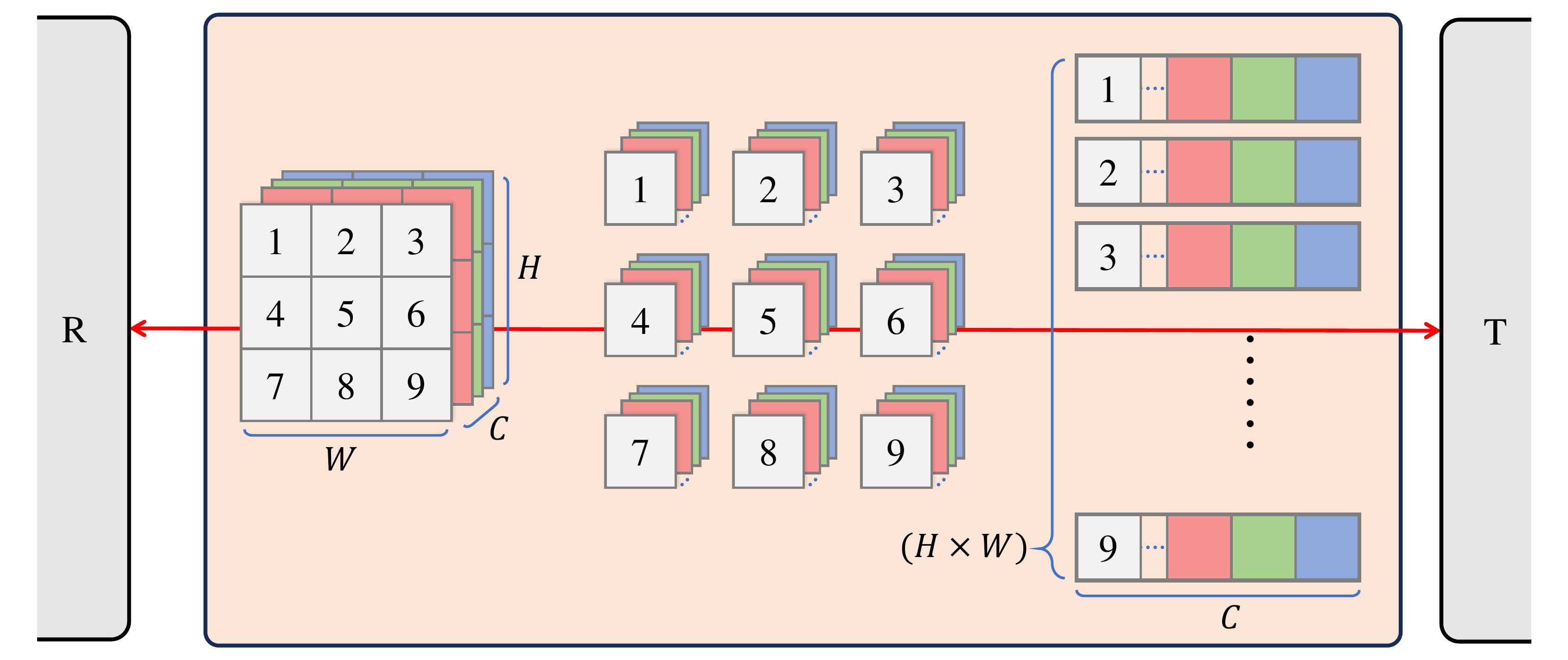}
        \label{fig:arch_conversion}
    }
    \captionsetup[subfigure]{justification=centering}
    \caption{ResTNet consists of a sequence of blocks, each of which is either a residual block (\texttt{R}) or a Transformer block (\texttt{T}). The feature conversion transfers the features between \texttt{R} and \texttt{T}. Subfigure (d) illustrates conversion from \texttt{R} to \texttt{T}. The one from \texttt{T} to \texttt{R} simply left-right mirrors the network. The ones from \texttt{R} to \texttt{R} and \texttt{T} to \texttt{T} use an identical mapping, i.e., no conversion.}
    \label{fig:architecture}
\end{figure*}

This paper proposes \textit{ResTNet}, a novel architecture that combines residual and Transformer blocks, as shown in Figure \ref{fig:architecture}.
It consists of several blocks, with each block being either a residual block or a Transformer block.
For any given board, the network outputs a policy distribution and a value, as depicted in Figure \ref{fig:arch_restnet}, where the architectures for the residual block and Transformer block are described in more detail in Figure \ref{fig:arch_r} and \ref{fig:arch_t} respectively.
\revision{The residual block follows the same architecture as AlphaZero \cite{silver_general_2018}, comprising two convolutional layers.
The Transformer block is based on the standard Transformer \cite{vaswani_attention_2017}, augmented with relative position encoding \cite{shaw_selfattention_2018}, and uses four attention heads per block.}

For simplicity, we use ``\texttt{R}'' to denote the residual block, and ``\texttt{T}'' to represent Transformer block in the network configuration for the rest of the paper.
For instance, \texttt{RRTRRT} represents a network configuration that begins with two residual blocks, followed by a Transformer block, then two additional residual blocks, and finally another Transformer block.
In addition to this notation, we employ a numerical naming method as an alternative way.
For example, \texttt{RRTRRT} corresponds to the same configuration as \texttt{2R1T2R1T}, while \texttt{6R1T} represents a network configuration starting with six consecutive residual blocks followed by one Transformer block.

ResTNet offers a general version that covers a variety of network architectures.
For example, in a 3-block configuration of ResTNet, if all blocks are residual blocks, represented as \texttt{RRR} or \texttt{3R}, this configuration matches the architecture of the AlphaZero network.
Conversely, if all blocks are Transformer blocks, denoted as \texttt{TTT} or \texttt{3T}, the network aligns with ViT.
Additionally, a configuration that starts with residual blocks and transitions to Transformer blocks towards the end, such as \texttt{RRT} or \texttt{RTT}, corresponds to the CoAtNet architecture.
Note that there are still some configurations, such as \texttt{TRR} or \texttt{RTR}, that are not covered in the above architectures but are included in ResTNet.

\subsection{Feature Conversion}
\label{feature_conversion}
In ResTNet, the residual and Transformer blocks process different input features.
The residual blocks utilize convolutional neural networks to process \textit{2D feature maps}, whereas the Transformer blocks, designed for sequences handling, use \textit{1D tokens} as input.
In board games, precise and accurate positional representation is crucial, as each position on the board is unique and can differ significantly from neighboring positions.
Therefore, we present a feature conversion method for transferring features between residual and Transformer blocks in ResTNet.
Figure \ref{fig:arch_conversion} illustrates feature conversion from residual to Transformer blocks. 
Note that the feature conversion from Transformer to residual blocks simply mirror Figure \ref{fig:arch_conversion} left-right. 
The method converts the 2D feature maps into 1D tokens through a one-to-one mapping to preserve positional information, as described as follows.

The input/output representations of the residual block are consistent with the board representation, which is defined by dimensions $C\times H\times W$, where $C$ represents the number of channels, and both $H$ and $W$ denote the height and width respectively.
Hence, the initial board representation can be viewed as a specification of the residual block to the first conversion. 
Conversely, for Transformer blocks, the representation needs to be converted into 1D tokens, ordered according to the board positions, from the top-left to the bottom-right in either row-major or column-major manner.
The size of 1D tokens is $H \times W$, with each token encoding a board position in a dimension of $C$.
For a conversion from Transformer to residual block, these 1D tokens can further be reorganized into 2D feature maps, ensuring that each token is accurately rearranged to match the original board layout.

\section{Experiments}
\label{sec:experiment}
This section presents an in-depth analysis of ResTNet in Go and Hex, selected for their reliance on global knowledge to understand gameplay.
Subsection \ref{exp:restnet_performance} evaluates the playing performance.
Subsection \ref{exp:global_info} examines ResTNet's ability to process global information, focusing on two well-known challenges in Go: recognizing \textit{circular patterns} and \textit{ladder patterns}.
Finally, Subsection \ref{exp:visualization_restnet} explores the attention maps in RestNet, offering insights into its explainability.

\subsection{Playing Performance of ResTNet}
\label{exp:restnet_performance}
Given that the proposed ResTNet allows for various combinations, we first evaluate different network architectures on 9x9 Go, which allows multiple evaluations under limited computational resources.
Then, we apply the best-performing architecture discovered to 19x19 Go and 19x19 Hex.

\subsubsection{Architecture Exploration on 9x9 Go}
\label{exp:restnet_9x9go}
We use 6-block in 9x9 Go.
Each residual block consists of 256 filters, whereas each Transformer block comprises 81 ($9\times9$) tokens with an embedding size of 256.
Table \ref{tab:simulation_paras} lists the 6-block networks we evaluated, along with their inference time and the number of parameters.
These networks can be further categorized into four types.
First, the AlphaZero-like networks, comprising solely of convolutional neural networks, i.e., \texttt{6R}.
Second, the ViT-like networks, consisting purely Transformer-based networks, i.e., \texttt{6T}.
Third, the CoAtNet-like networks begin with a series of residual blocks followed by Transformer blocks, such as \texttt{5R1T}, \texttt{4R2T}, etc.
Finally, the networks with interleavings of residual and Transformer blocks, such as \texttt{RTRRRT}, \texttt{RRTRRT}, etc.
Considering the inference time, we only explore permutations that include two Transformer blocks and four residual blocks, with one Transformer block at the end to facilitate global information extraction\footnote{We also try placing R at the end, but it leads to worse results, similar to the finding in computer vision.}.
Note that these four categories are all encompassed within our proposed ResTNet.

\begin{table} [t]
    \centering
    \setlength{\tabcolsep}{1mm}
    \resizebox{0.99\columnwidth}{!}{
    \begin{tabular}{clccc}
        \toprule
        \  & \textbf{Models} & \textbf{Time} & \textbf{Parameters}& \textbf{Win Rate} \\
        \midrule
        \ \textbf{Conv only} & \texttt{6R} & 3.067 & 7.146 & 54.60\% $\pm$ 2.15\%\\
        \midrule
        \ \textbf{ViT-like} & \texttt{6T} & 8.130 & 3.358 & 39.85\% $\pm$ 2.18\%\\
        \midrule
        \ \multirow{5}{*}{\textbf{CoAtNet-like}} & \texttt{5R1T} & 3.650 & 6.619 & 56.00\% $\pm$ 2.18\%\\
        \  & \texttt{4R2T} & 4.237 & 5.967 & 51.75\% $\pm$ 2.19\%\\
        \  & \texttt{3R3T} & 5.000 & 5.314 & 47.85\% $\pm$ 2.19\%\\
        \  & \texttt{2R4T} & 5.882 & 4.662 & 37.40\% $\pm$ 2.12\%\\
        \  & \texttt{1R5T} & 6.329 & 4.001 & 31.15\% $\pm$ 2.03\%\\
        \midrule
        \ \multirow{4}{*}{} & \texttt{\textbf{T}RRRR\textbf{T}} & 4.566 & 5.967 & 43.90\% $\pm$ 2.18\%\\
        \  & \texttt{R\textbf{T}RRR\textbf{T}} & 4.566 & 5.967 & 54.35\% $\pm$ 2.18\%\\
        \  & \texttt{RR\textbf{T}RR\textbf{T}} & 4.566 & 5.967 & \textbf{60.80}\% $\pm$ \textbf{2.14}\% \\
        \  & \texttt{RRR\textbf{T}R\textbf{T}} & 4.566 & 5.967 & 49.10\% $\pm$ 2.19\%\\
        \bottomrule
    \end{tabular}
    }
        \caption{Various 6-block ResTNet model configurations, including inference time (in milliseconds), the number of parameters (in millions), and the win rate against KataGo under 2 seconds thinking time, along with the 95\% confidence interval.}
    \label{tab:simulation_paras}
\end{table}

Unlike in computer vision, where accuracy is commonly used as a benchmark to measure the performance of networks, our evaluation focuses on the playing strength.
Specifically, we train each network using the Gumbel AlphaZero algorithm \cite{danihelka_policy_2022} with 64 simulations based on an open-sourced AlphaZero framework \cite{wu_minizero_2024}.
Each training generates a total of 1 million self-play games and includes 100,000 network optimization steps, requiring approximately 200 1080Ti GPU hours.
\revision{For the evaluation, we compare our models against four KataGo models\footnote{These models are kata1-b6c96-s115648256, kata1-b10c128-s41138688, kata1-b10c128-s108710656, and kata1-b15c192-s798345984.} as reference opponents across a total of 2,000 games.}

From Table \ref{tab:simulation_paras}, models with an equal or greater number of Transformer blocks than residual blocks, i.e., \texttt{6T}, \texttt{1R5T}, \texttt{2R4T}, and \texttt{3R3T}, obtain a win rate of under 50\% against KataGo.
This is likely due to the lack of convolutional layers which are crucial for capturing local patterns in board games.
For CoAtNet-like models, as the number of consecutive Transformer blocks increases, we observe a corresponding decrease in strength.
This result suggests that using a long sequence of Transformer blocks does not necessarily improve performance.
Surprisingly, this contradicts findings from CoAtNet in computer vision, where using consecutive Transformer blocks is suggested to yield better results.
For networks that interleave residual and Transformer blocks, starting with Transformer blocks, i.e., \texttt{TRRRRT} performs the worst.
This indicates the importance of initially extracting local patterns before processing global patterns.
On the other hand, \texttt{RRTRRT} achieves the highest win rate of 60.80\% against KataGo among all models.
We conjecture that this repeating pattern of \texttt{RRT} effectively balances the extraction of global information while simultaneously preserving local patterns.

\subsubsection{Performance on 19x19 Go and 19x19 Hex}
\label{exp:restnet_19x19go_19x19hex}
We further evaluate the playing performance on 19x19 Go and 19x19 Hex based on the experiments from the previous subsection.
We select two 10-block network models: \texttt{10R} and \texttt{R3(RRT)}.
\texttt{10R} serves as an AlphaZero-like baseline model, while \texttt{R3(RRT)}, also denoted as \texttt{RRRTRRTRRT}, begins with \texttt{R} and follows a repeating sequence of three \texttt{RRT}.

For 19x19 Go, to reduce the computational costs of training models from scratch, we trained these models using supervised learning on a human game collection.
\revision{The collection contains a total of 1 million games played by 7 dan to 9 dan human Go players on Tygem~\cite{cho_tygemgo_2001}, a popular online Go platform.}
\revision{After training both models to a total of 150,000 optimization steps, we evaluate their performance against four KataGo models\footnote{These models are kata1-b10c128-s41138688, kata1-b10c128-s108710656, kata1-b10c128-s46989824, and kata1-b10c128-s56992512.} under the same thinking time of 5 seconds per move across a total of 2,000 games.}
For 19x19 Hex, we train two 10-block models directly using the \revision{Gumbel} AlphaZero algorithm \revision{with 32 simulations}.
\revision{Each model trains with 500,000 self-play games and 100,000 optimization steps.}
\revision{Then, we evaluate the playing performance of the models by playing 1,000 games against MoHex \cite{arneson_monte_2010}, a well-known Hex program that won championships in computer Olympiads.}

The results in Table \ref{tab:19b_model_eval} show that \texttt{R3(RRT)} outperforms \texttt{10R} in both 19x19 Go and 19x19 Hex.
In summary, these experiments demonstrate that the discovered architecture, repeating \texttt{RRT}, efficiently integrates Transformers with residual blocks and improves playing performance in board games.

\begin{table}[t]
\centering
\begin{small}
\begin{tabular}{lrr}
\toprule
 & \textbf{19x19 Go} & \textbf{19x19 Hex} \\
\midrule
\texttt{10R} & 53.60\% $\pm$ 2.19\% & 50.40\% $\pm$ 4.39\% \\
\texttt{R3(RRT)} & \textbf{60.90}\% $\pm$ \textbf{2.14}\% & \textbf{58.00}\% $\pm$ \textbf{4.33}\% \\
\bottomrule
\end{tabular}%
\end{small}
\caption{The playing performance of \texttt{10R} and \texttt{R3(RRT)} in 19x19 Go and 19x19 Hex respectively.
The results include 95\% confidence intervals.
}
\label{tab:19b_model_eval}
\end{table}

\subsection{Global Information Ability of ResTNet}
\label{exp:global_info}
This subsection investigates the ability of the two trained 19x19 Go models, \texttt{10R} and \texttt{R3(RRT)}, as described in the previous subsection, to process global information by evaluating their performance on two well-known challenges in Go: recognizing \textit{circular patterns} and \textit{ladder patterns}.

\subsubsection{Recognition of Circular Patterns}
\label{exp:circular_patterns}
Circular patterns refer to long sequences of blocks that form a circular enclosure around the opponent's stones, usually spanning a wide area of the Go board, as shown in the top-right corner of Figure \ref{fig:intro_circular}.
These patterns present a specific challenge for Go programs that require effective global information processing.
\citep{wang_adversarial_2023} developed an adversarial Go program called \textit{cyclic-adversary}, which effectively induces circular patterns, causing most Go programs to misjudge their life-and-death status due to an insufficient understanding of global information.
In addition, \citep{wang_adversarial_2023} provided a game collection containing 24 games, each featuring circular patterns, played by cyclic-adversary against KataGo.

\begin{figure}[t]
\centering
\subfloat[]{\includegraphics[width=0.48\columnwidth]{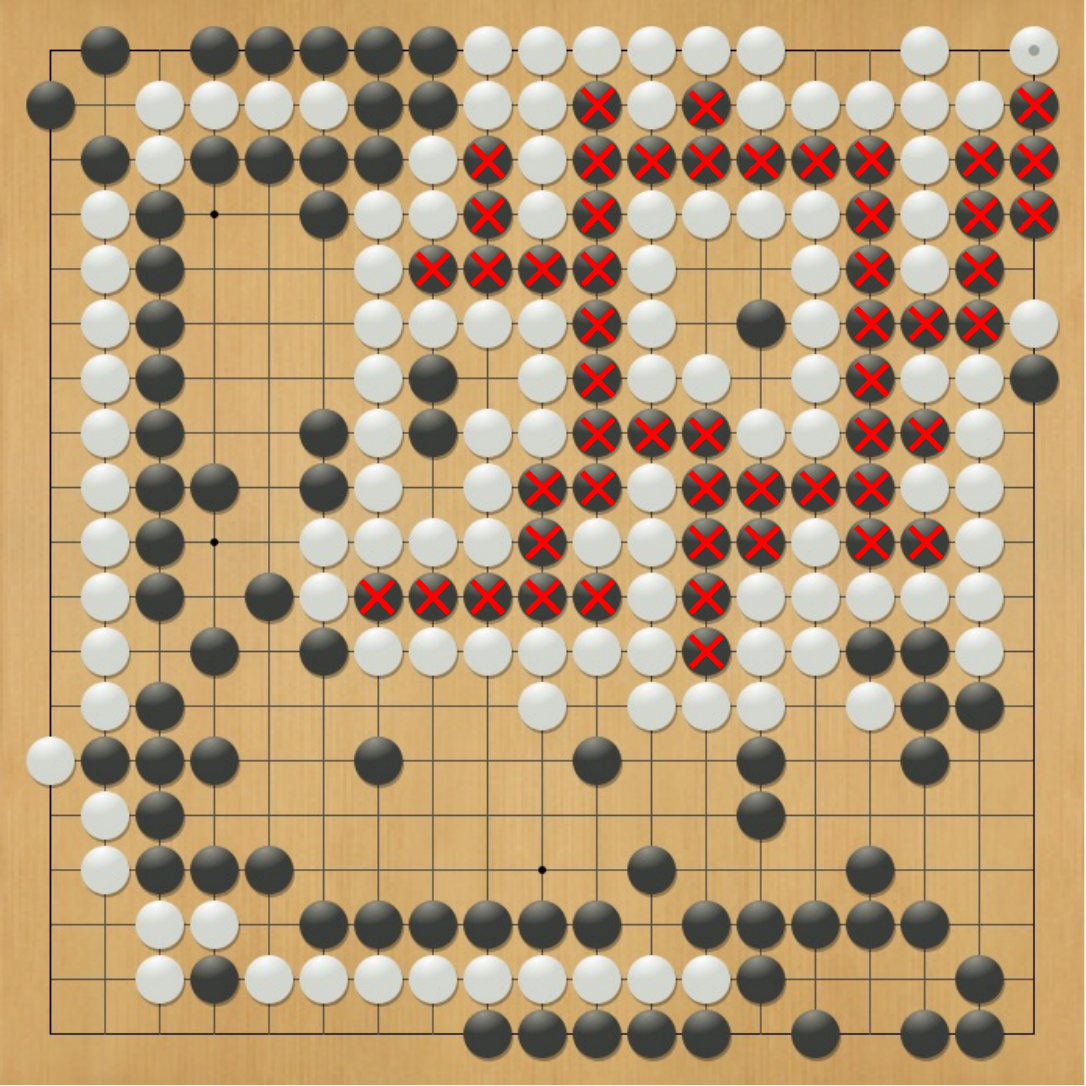}
\label{fig:circular_r1_a}
}
\subfloat[Ground truth]{\includegraphics[width=0.48\columnwidth]{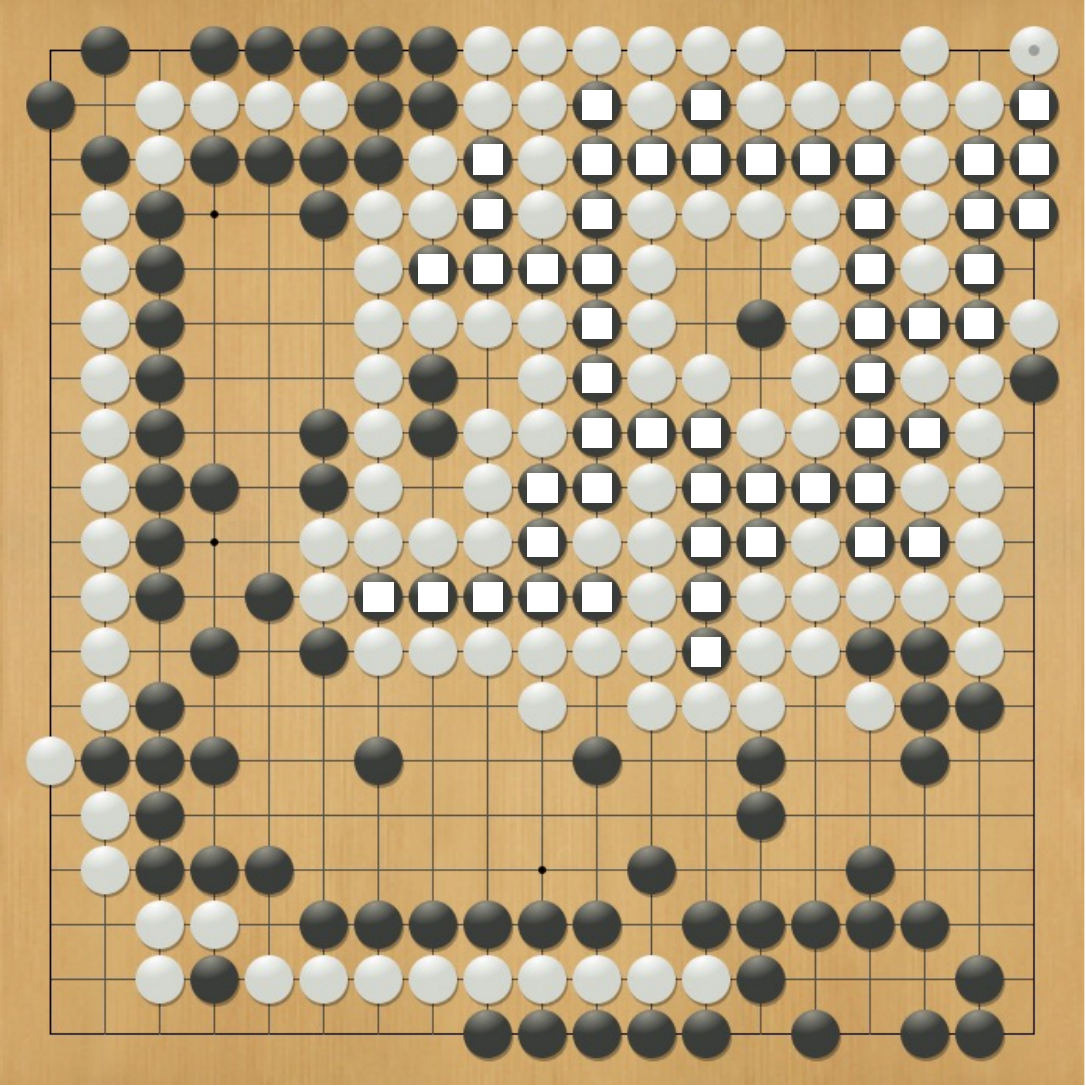}
\label{fig:circular_r1_b}
}
\\
\subfloat[\texttt{10R}]{\includegraphics[width=0.48\columnwidth]{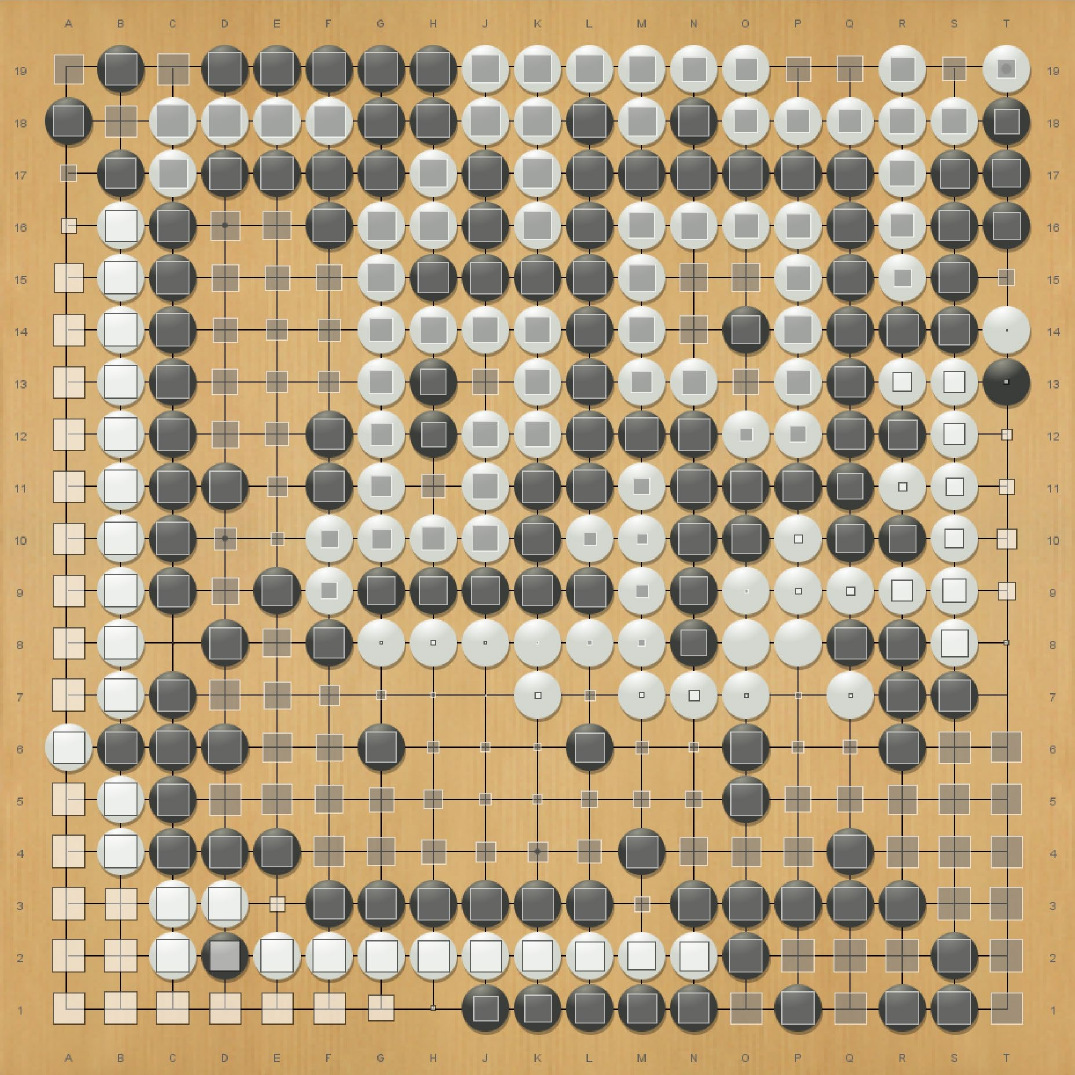}
\label{fig:circular_r1_c}
}
\subfloat[\texttt{R3(RRT)}]{\includegraphics[width=0.48\columnwidth]{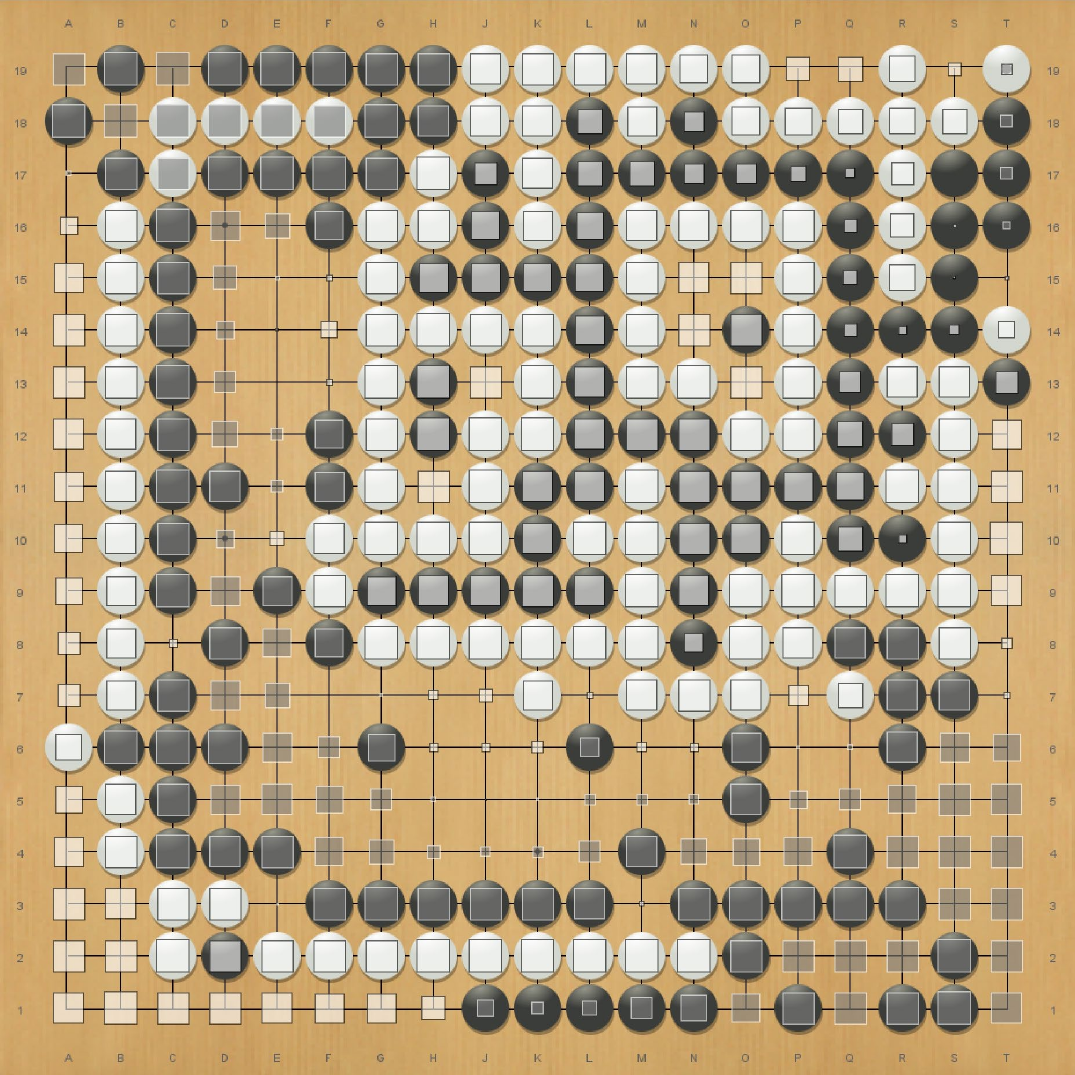}
\label{fig:circular_r1_d}
}
\caption{The circular pattern, ground truth results, and board evaluation results for \texttt{10R} and \texttt{R3(RRT)}.
\revision{The square in (c) and (d) indicates the ownership predicted by each model, with larger sizes representing higher confidence in ownership by black or white.}}
\label{fig:circular_r1}
\end{figure}

\revision{We first evaluate whether ResTNet can correctly recognize the life-and-death status of circular patterns in the game collection\footnote{\label{fn:wangs}These games are available at \url{https://goattack.far.ai/adversarial-policy-katago}.} provided by \citep{wang_adversarial_2023}.}
Specifically, we incorporate a board evaluation head \cite{gilmer_jmgilmer_2016,wu_multilabeled_2018} into ResTNet.
The board evaluation head, commonly used in current Go programs to assess a model's understanding of life-and-death status, is designed to predict the ownership of each position on the board at the endgame.
The output values are bounded within the range of [-1, 1], where 1 represents black ownership and -1 represents white ownership.
Figure \ref{fig:circular_r1_a} shows an example of circular pattern games from the game collection.
In this game, the cyclic-adversary, playing as White, successfully establishes a circular pattern and deceives KataGo, resulting in the capture of the marked black stones.
The ownership of these marked stones should be white, as shown in \ref{fig:circular_r1_b}.
Interestingly, significant differences are observed between the evaluations of \texttt{10R} and \texttt{R3(RRT)}, as depicted in Figure \ref{fig:circular_r1_c} and \ref{fig:circular_r1_d}.
The \texttt{10R} incorrectly predicts the marked stones as belonging to the black player.
In contrast, the ownership predicted by \texttt{R3(RRT)} accurately aligns with the ground truth, demonstrating robustness in recognizing circular patterns.
We evaluate all games and quantify model performance using the mean square error (MSE) between the ground truth and board evaluation output, as summarized in Table \ref{tab:cyclic_attack_rate_bv_mse_pred_ladder_ptrn}.
The results show that \texttt{R3(RRT)} achieves a significantly lower MSE compared to \texttt{10R}, indicating a more accurate understanding of the life-and-death status for circular patterns.

Moreover, since \texttt{R3(RRT)} demonstrates a better understanding of circular patterns, it is worth investigating whether \texttt{R3(RRT)} can defend against the cyclic-adversary.
\revision{Specifically, both \texttt{R3(RRT)} and \texttt{10R} play against the cyclic-adversary using the same openings from \citep{wang_adversarial_2023}'s game collection, with each opening beginning with a circular pattern.}
The model must recognize circular patterns and play a correct sequence of moves to avoid being captured and win the game; otherwise, it will lose.
Given the inherent randomness of the cyclic-adversary, each opening is played by 30 games, resulting in a total of 720 evaluation games.
Table \ref{tab:cyclic_attack_rate_bv_mse_pred_ladder_ptrn} shows the probability of each model being attacked by the cyclic-adversary.
Compared to \texttt{10R}, \texttt{R3(RRT)} significantly reduces the being attacked rate from 70.44\% to only 23.91\%, a reduction by a factor of 2.95.
Overall, the results demonstrate that using ResTNet enhances the ability to leverage global knowledge for processing long sequences, addressing challenging problems that most Go programs struggle to handle.

\begin{table}[t]
\centering
\resizebox{0.99\columnwidth}{!}{
\begin{tabular}{l|r@{\hspace{0em}}cc@{\hspace{0em}}c|r@{\hspace{0em}}r}
\toprule
\multirow{2}{*}{} && \multicolumn{2}{c}{\textbf{Circular Patterns}} &&& \textbf{Ladder Patterns} \\
\cline{3-4} \cline{7-7}
&& \makecell[c]{\raisebox{-.5ex}{MSE}} & \makecell[c]{\raisebox{-.5ex}{Being Attacked Rate}} &&& \makecell[c]{\raisebox{-.5ex}{Accuracy}} \\
\midrule
\texttt{10R} && 2.58 $\pm$ 0.59 & 70.44\% $\pm$ 3.34\% &&& 59.15\% $\pm$ 0.24\% \\
\texttt{R3(RRT)} && \textbf{1.07 $\pm$ 0.37} & \textbf{23.91}\% $\pm$ \textbf{3.13}\% &&& \textbf{80.01}\% $\pm$ \textbf{0.19}\% \\
\bottomrule
\end{tabular}%
}
\caption{Experiments for ResTNet's global information ability on long-sequence patterns. For circular patterns, the MSE measures the error in recognizing the life-and-death status of circular patterns, while the being attacked rate represents the probability of being attacked by the cyclic-adversary. Lower values are better for both metrics. For ladder patterns, the accuracy indicates the model's ability to correctly predict ladder pattern outcomes. All results include 95\% confidence intervals.
}
\label{tab:cyclic_attack_rate_bv_mse_pred_ladder_ptrn}
\end{table}

\subsubsection{Recognition of Ladder Patterns}
\label{exp:ladder_patterns}
\begin{figure}[t]
\centering
\subfloat[Escape success]{\includegraphics[height=0.48\columnwidth]{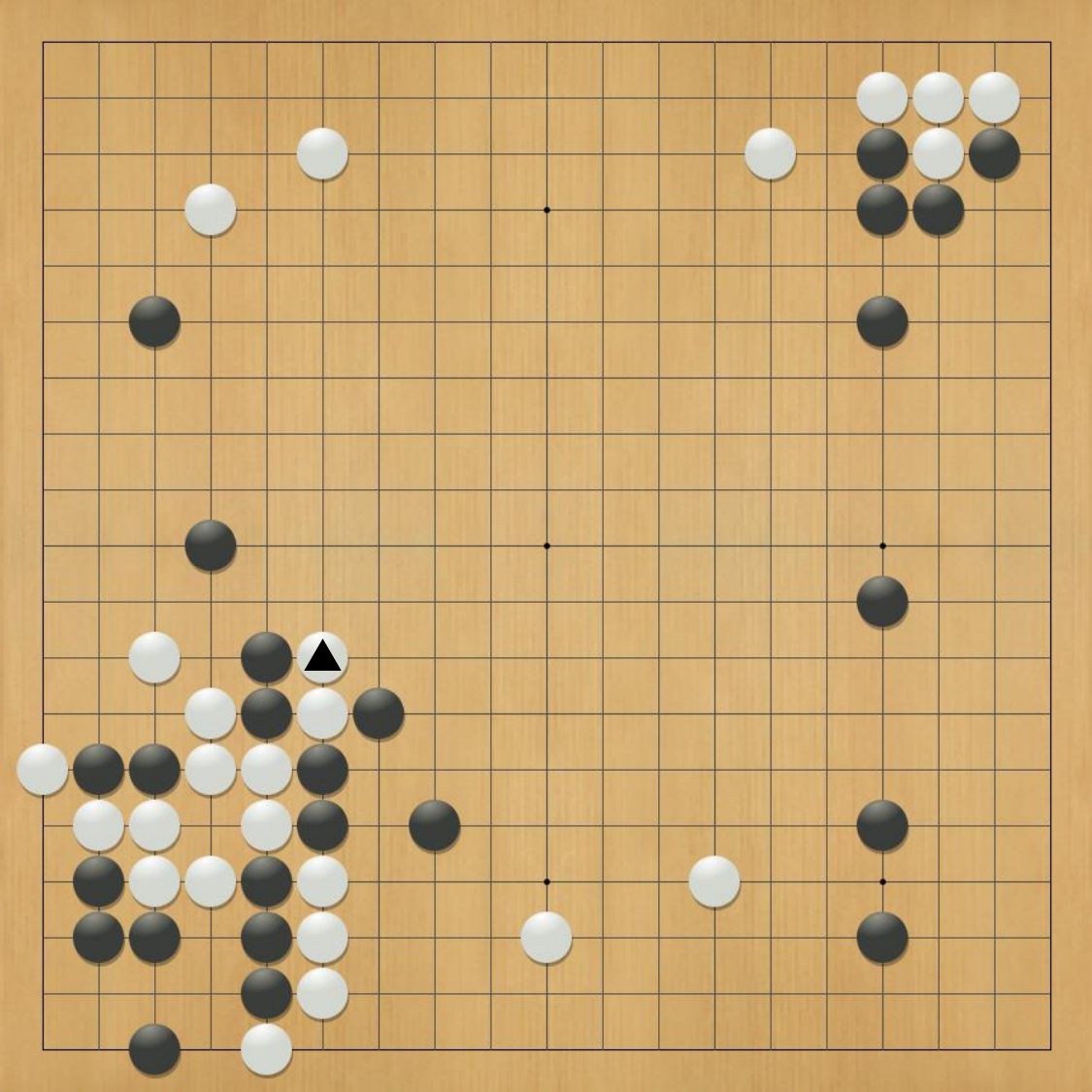}
\label{fig:ladder_escape_success}
}
\subfloat[Success result]{\includegraphics[height=0.48\columnwidth]{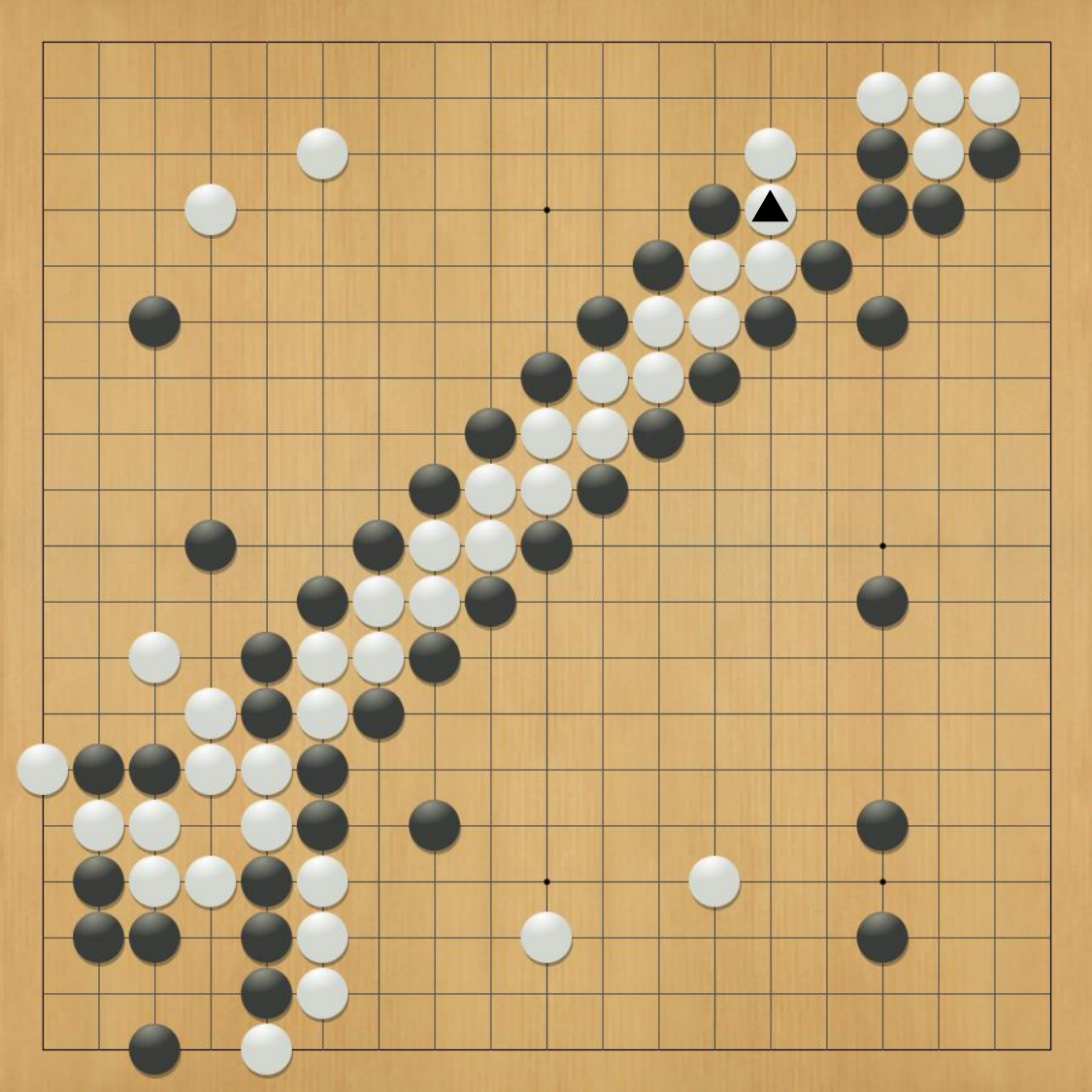}
\label{fig:ladder_escape_success_result}
}
\\
\subfloat[Escape failure]{\includegraphics[height=0.48\columnwidth]{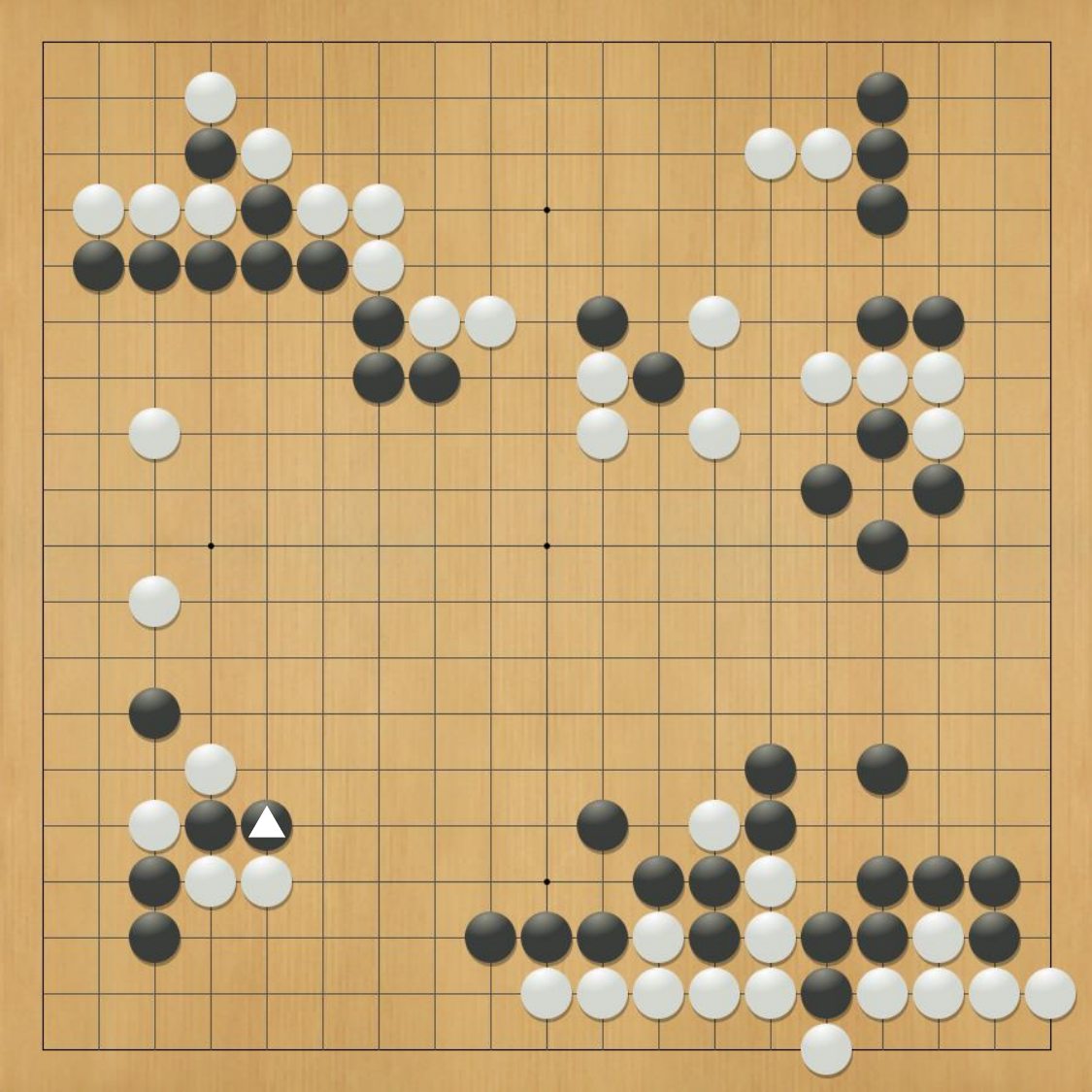}
\label{fig:ladder_escape_failure}
}
\subfloat[Failure result]{\includegraphics[height=0.48\columnwidth]{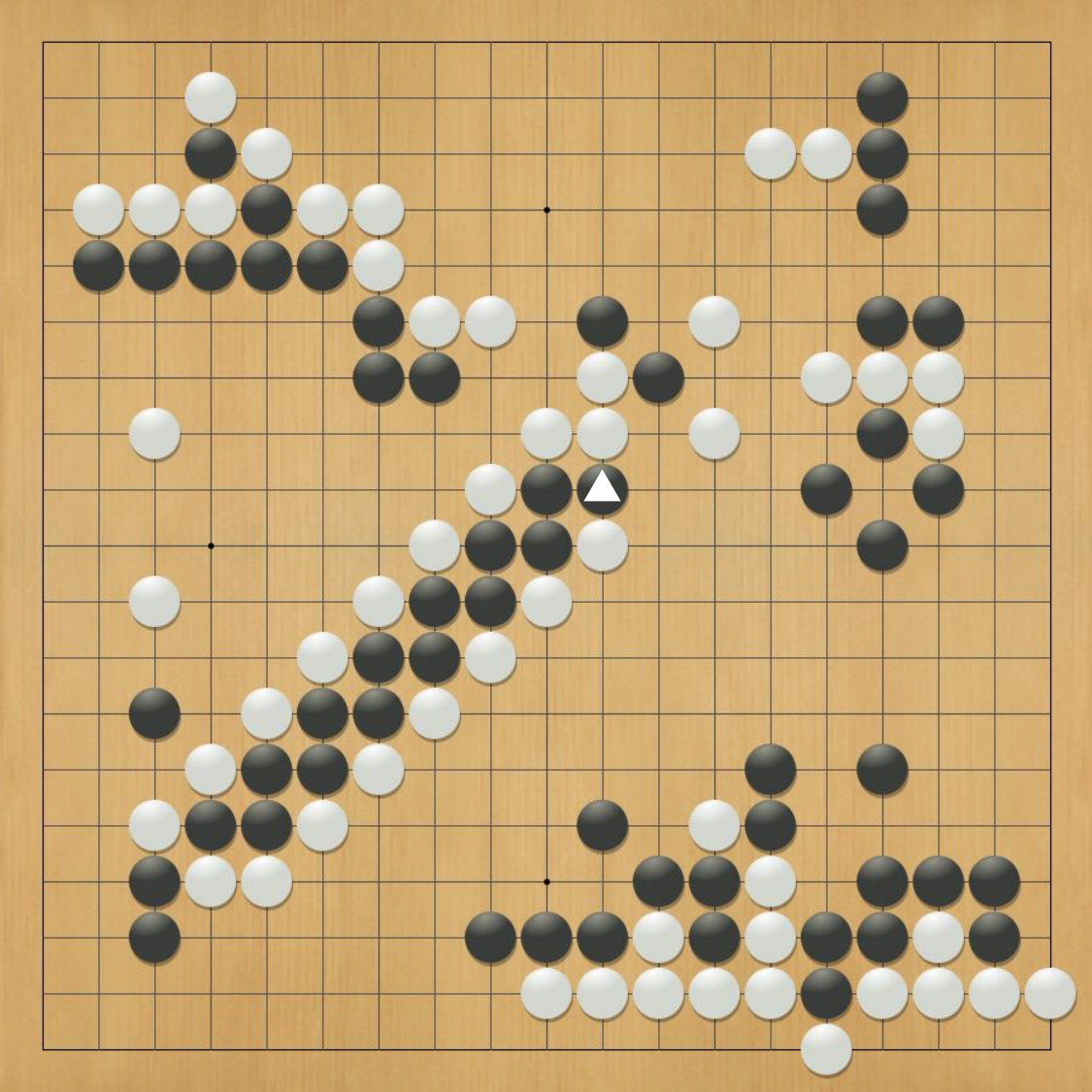}
\label{fig:ladder_escape_failure_result}
}
\caption{Two ladder patterns from the ladder dataset. (a) and (c) are the original boards, while (b) and (d) show the respective solutions.}
\label{fig:ladder_examples_d}
\end{figure}

We investigate whether ResTNet can effectively address another long sequence challenge -- recognizing the \textit{ladder pattern} in 19x19 Go.
Unlike circular patterns, which can be generated using cyclic-adversary, there are no established strategies to induce ladder patterns.
Therefore, we collected a total of 1,655,000 ladder patterns from the Tygem game collection.
Figure \ref{fig:ladder_escape_success} and \ref{fig:ladder_escape_failure} show two examples of ladder patterns.
The stones marked with triangles indicate the player attempting to escape, serving as the defender, while the opponent is the attacker, trying to capture the marked stone.
Note that the ladder dataset includes an equal number of escape successes and escape failures to ensure fairness.

Next, we use model probing \cite{alain_understanding_2017} to examine whether the trained models capture the ladder patterns.
Specifically, we incorporate a \textit{ladder head} into the network output, freezing the backbone parameters while allowing only the ladder head to be trained.
The ladder head aims to predict whether the defender can escape: 1 for success and -1 for failure, as shown in Figure \ref{fig:ladder_escape_success_result} and \ref{fig:ladder_escape_failure_result}.
This provides a straightforward approach to determine whether the backbone has learned ladder-related information, as the ladder head cannot recognize ladder patterns without this information.
During the evaluation, for each ladder pattern, the model predicts escape success if the output of the ladder head is greater than 0.5, and escape failure if the output is less than -0.5.
Values within the range of $(-0.5, 0.5)$ are classified as unknown and are considered incorrect predictions.

We train the ladder head on both \texttt{10R} and \texttt{R3(RRT)}.
After training, we evaluate these models on a separate testing ladder dataset consisting of 166,500 ladder patterns.
Table \ref{tab:cyclic_attack_rate_bv_mse_pred_ladder_ptrn} shows the accuracy of \texttt{10R} and \texttt{R3(RRT)} in identifying ladder patterns, where \texttt{R3(RRT)} demonstrates a significant improvement, increasing accuracy from 59.15\% to 80.01\%.
This aligns with the findings from ELF OpenGo \cite{tian_elf_2019}, which indicate that AlphaZero-like Go programs struggle to fully recognize ladder patterns.
In contrast, the inclusion of Transformer blocks significantly enhances the model's ability to process long sequences and recognize ladder patterns.

\subsection{Visualization of ResTNet}
\label{exp:visualization_restnet}
We visualize ResTNet by constructing attention maps using the attention values from the Transformer block, where each token corresponds to an attention map, and the values represent the relative importance of other tokens.
By analyzing these attention maps, we can explore patterns or strategies utilized by ResTNet, offering insights into its behavior and decision-making process.
This analysis is conducted on both 19x19 Go and 19x19 Hex using \texttt{R3(RRT)}, the same models used in subsection \ref{exp:restnet_19x19go_19x19hex}.

\subsubsection{Attention Maps in 19x19 Go}
\label{exp:attenion_map_19x19go}
\begin{figure}[t]
\centering
\hspace{-0.45em}
\subfloat[Alive Stones]{
\includegraphics[height=0.295\columnwidth]{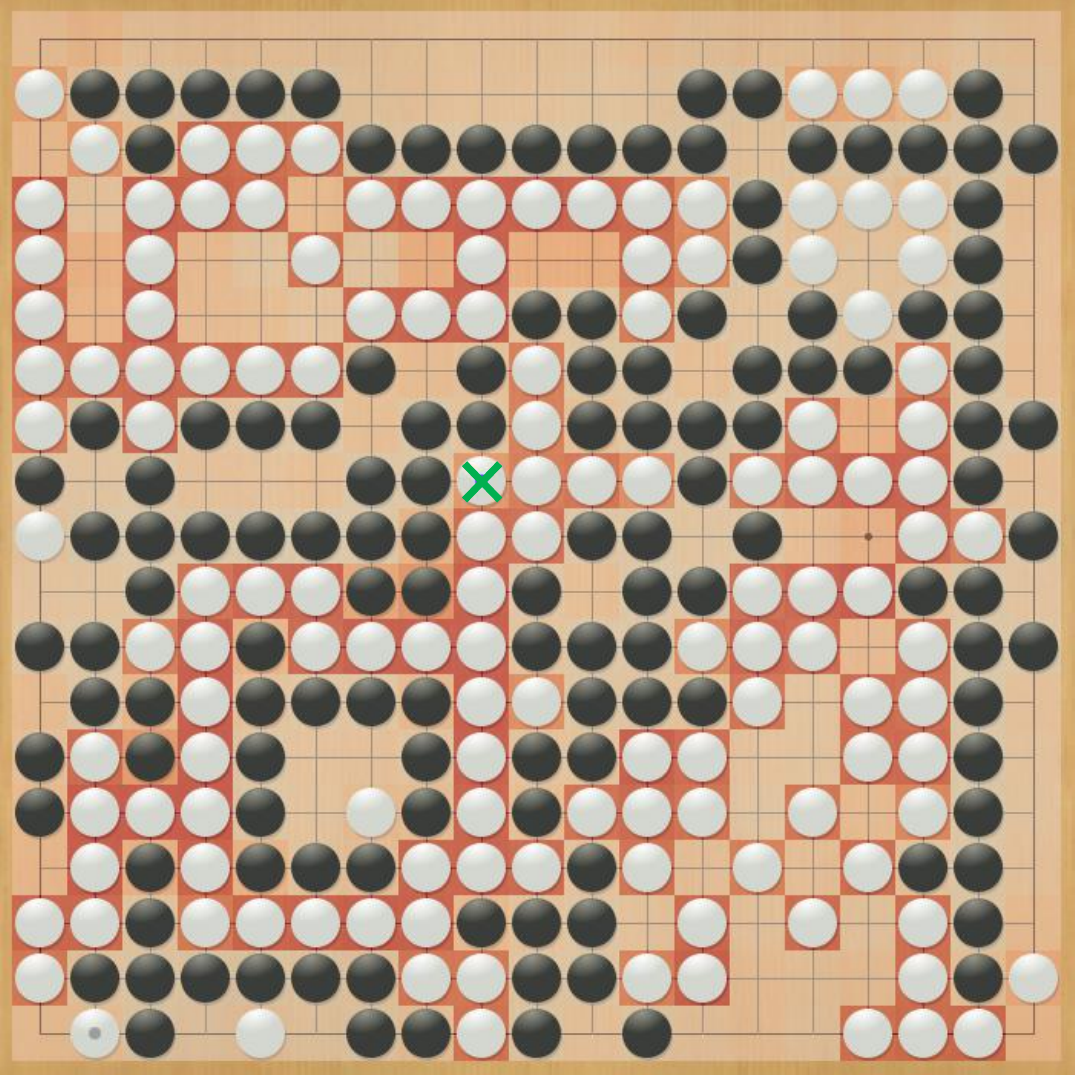}
\label{fig:attn_stone_pos}
}
\hspace{-0.42em}
\subfloat[Uncertain territory]{
\includegraphics[height=0.295\columnwidth]{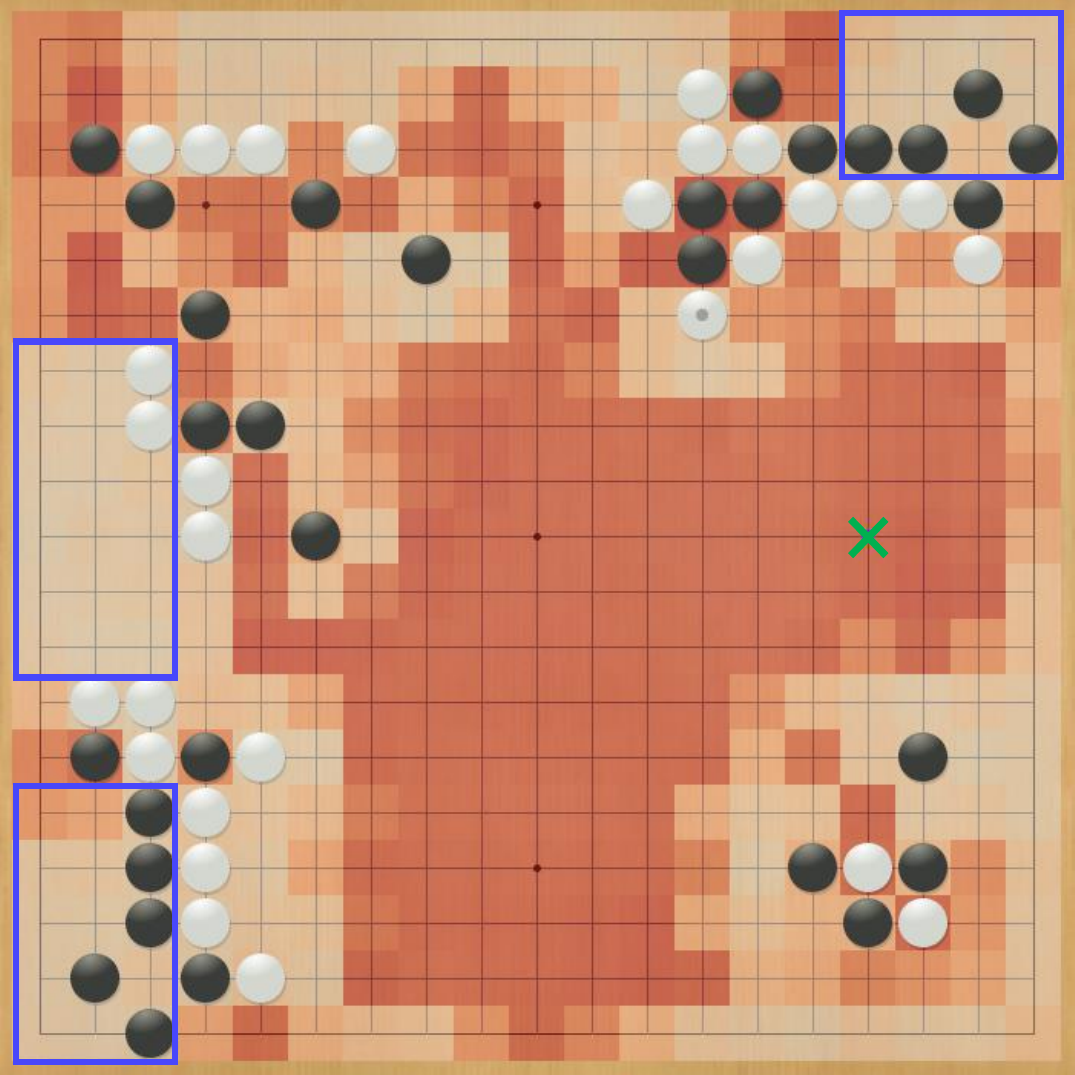}
\label{fig:attn_unconfirm_area}
}
\hspace{-0.42em}
\subfloat[Critical positions]{
\includegraphics[height=0.295\columnwidth]{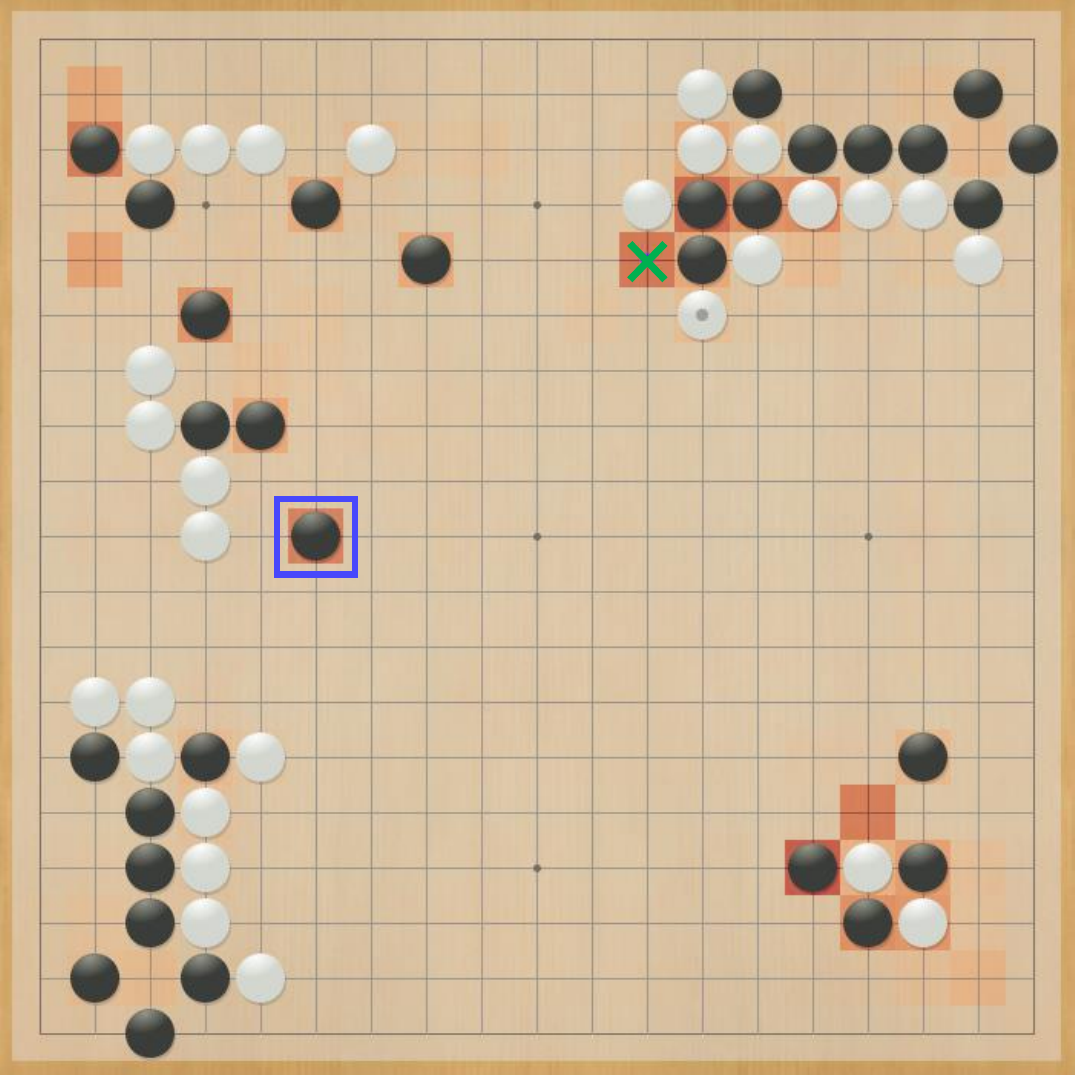}
\label{fig:attn_ladder_important_area}
}
\hspace{-0.55em}
\includegraphics[height=0.27\columnwidth]{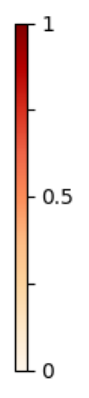}
\caption{
\revision{Since different blocks and heads may capture different types of knowledge, the attention values shown corresponding to the position marked in green, taken from the fourth head of the first \texttt{T} block for (a), the fourth head of the second \texttt{T} block for (b), and the third head of the second \texttt{T} block for (c).}
All positions are Black's turn to move. We normalize the values to [0, 1] for better visualization, with redder colors indicating higher values.} 
\label{fig:attention_maps}
\end{figure}

We select two 19x19 Go games, including a circular pattern and a ladder pattern.
Figure \ref{fig:attention_maps} illustrates the attention maps for the position marked in green, with redder colors indicating higher levels of relative importance.
Interestingly, these attention maps correspond closely to Go knowledge concepts.
First, the attention map in Figure \ref{fig:attn_stone_pos} focuses exclusively on the white stones that remain alive until the end game, aligning with the life-and-death concept in Go. 
Notably, these stones are spread across the entire board, making them extremely challenging for convolutional networks to recognize effectively.
Second, the attention map in Figure \ref{fig:attn_unconfirm_area} highlights uncertain territory.
The center area, not surrounded by any player, shows high attention values, while the three areas marked by blue rectangles, surrounded by one player, show low attention values.
Finally, the attention map in Figure \ref{fig:attn_ladder_important_area} highlights important positions.
The three black stones can escape by following a sequence of moves starting with the green-marked position and connecting to the middle left black stone marked by the blue square.
Interestingly, the attention map focuses on three black stones as well as the middle left black stone, indicating that it captures the concept of the ladder pattern.

\subsubsection{Attention Maps in 19x19 Hex}
\label{exp:attenion_map_19x19hex}
\begin{figure}[t]
\centering
\subfloat[Potential paths]{
\includegraphics[height=0.28\columnwidth]{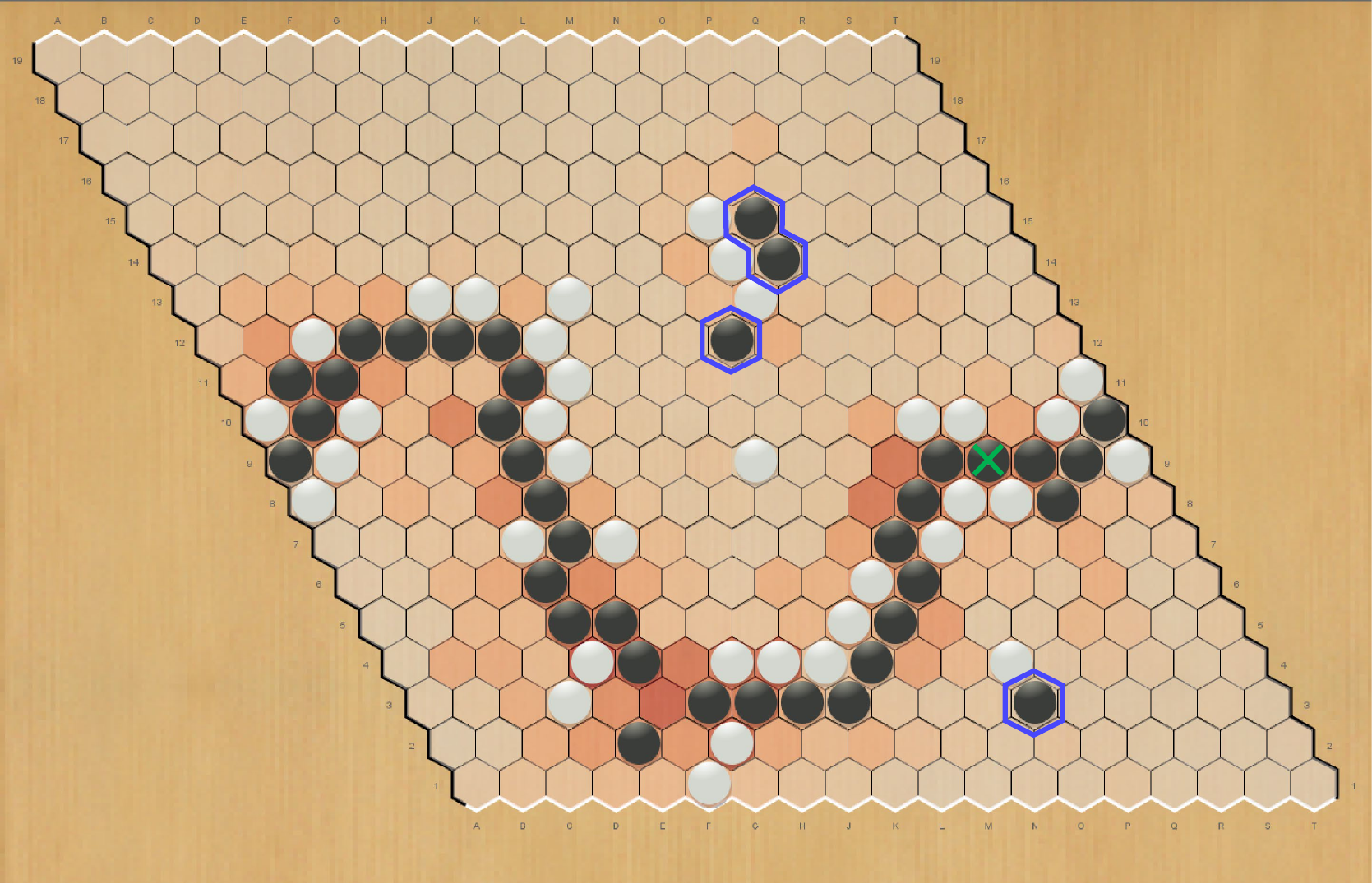}
\label{fig:attn_global_info}
}
\subfloat[Virtual connections]{
\includegraphics[height=0.28\columnwidth]{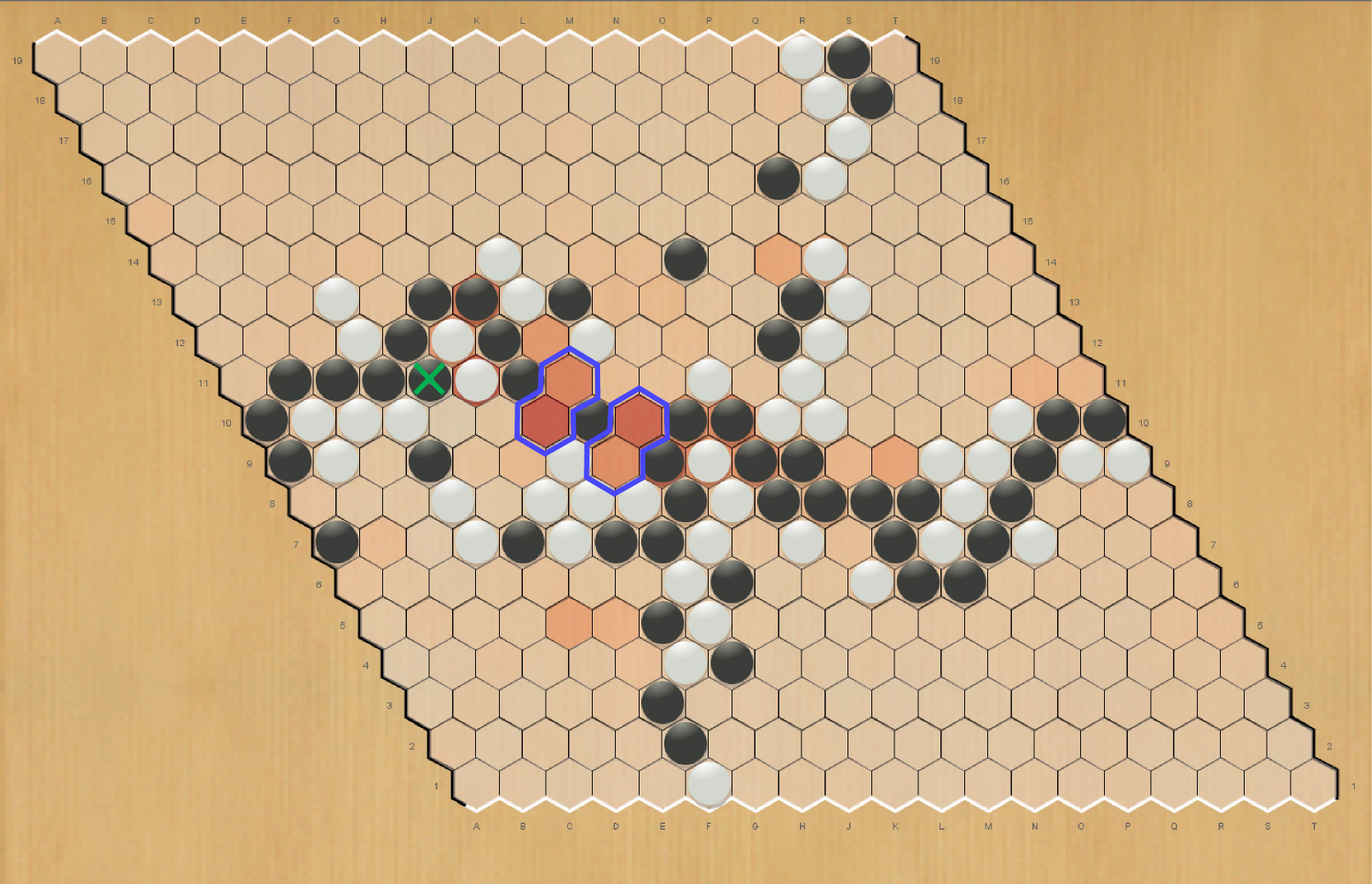}
\label{fig:attn_virtual_connection}
}
\includegraphics[height=0.28\columnwidth]{figures/Experiment/Attention_value/colorbar.pdf}
\caption{The attention maps for 19x19 Hex from \revision{the third head of} the second and the third \texttt{T} block for (a) and (b), respectively. The next moves of (a) and (b) are for White and Black, respectively.} 
\label{fig:hex_attention_maps}
\end{figure}

Next, we examine the attention maps in 19x19 Hex, as shown in Figure \ref{fig:hex_attention_maps}.
Similar to 19x19 Go, the attention maps in Hex show essential game concepts, particularly the fundamental strategy of connection.
The attention map in Figure \ref{fig:attn_global_info} highlights the concept of basic connection, focusing only on the black stones necessary to secure a win for Black, while irrelevant black stones, marked by blue hexagons, are excluded.
Identifying connection stones across the entire board from left to right also demonstrates that ResTNet effectively captures global information.
Moreover, in Figure \ref{fig:attn_virtual_connection}, the attention map shows a critical and advanced strategy in Hex, known as \textit{virtual connections} \cite{hayward_solving_2004}, which refer to sets of positions that enable one player to connect two specified sets of stones, even if the opponent plays first.
In conclusion, the attention maps in both Go and Hex demonstrate that ResTNet effectively learns global knowledge concepts while also offering opportunities for interpretability and explainability in board games.

\section{Discussion}
\label{sec:discussion}
This paper proposes \textit{ResTNet}, a network architecture specifically designed for AlphaZero algorithms in board games, integrating residual and Transformer blocks to bridge local and global knowledge.
Our experiments show that repeating \texttt{RRT} blocks achieves the best performance, effectively balancing inference time and model parameters.
ResTNet not only improves playing strength in both Go and Hex games, but also significantly addresses two challenging long-sequence patterns in 19x19 Go.
For circular patterns, ResTNet reduces the mean square error from 2.58 to 1.07 and lowers the average probability of being attacked against cyclic-adversary from 70.44\% to 23.91\%, significantly reducing a factor of 2.95.
For ladder patterns, ResTNet improves the accuracy of correctly predicting the outcome of ladder patterns from 59.15\% to 80.01\%.
Furthermore, visualizing attention maps reveals that ResTNet captures game-specific strategies, further contributing to the interpretability of AlphaZero.

\revision{
Due to computational constraints, this paper mainly explores ResTNet with repeating \texttt{RRT} in the main experiments.
However, even without a large-scale network architecture search, \texttt{RRT} still achieves superior performance.
Future work can build on this direction using larger-scale architecture search to further optimize ResTNet.
Moreover, as ResTNet is game-independent, it can be seamlessly applied to other board games, such as Chess or Shogi, without requiring modifications.}
Future research could explore its robustness by designing other adversarial strategies to further improve its capabilities.
While current board game programs have generally achieved superhuman performance, understanding their decision-making processes remains a critical challenge.
ResTNet's attention maps open promising research directions in explainable AI (XAI) for board games.
In conclusion, we believe the proposed ResTNet and our empirical findings make a valuable contribution to the board game community, offering a strong foundation for future research in this field.

\section*{Acknowledgments}

\revision{This research is partially supported by the National Science and Technology Council (NSTC) of
the Republic of China (Taiwan) under Grant Number NSTC 113-2221-E-001-009-MY3, NSTC
113-2634-F-A49-004, and NSTC 113-2221-E-A49-127.
The authors would also like to thank anonymous reviewers for their valuable comments.}


\bibliography{ijcai25}
\bibliographystyle{named}

\newpage



\section{Experiment Details for Training ResTNet}
\label{appendix:training_details}

This section details the training settings for the models used in the experiments described in Subsection 4.1,``Playing Performance of ResTNet," of the main text.
The training settings and model architectures for 9x9 Go, 19x19 Go, and 19x19 Hex are summarized in Table \ref{tab:training_setting}.

\begin{table}[h]
    \centering
    \resizebox{\columnwidth}{!}{%
    \begin{tabular}{llrrr}
        \toprule
        \textbf{Game} & & \textbf{9x9 Go} & \textbf{19x19 Go} & \textbf{19x19 Hex}\\
        \midrule
        \multirow{4}{*}{\textbf{Network Architecture}} & Number of blocks & 6 & 10 & 10 \\
        & Hidden channels & 256 & 256 & 256 \\
        & Transformer heads & 4 & 4 & 4 \\
        & Transformer MLP ratio & 2 & 2 & 2 \\
        \midrule
        \multirow{2}{*}{\textbf{Gumbel AlphaZero}} 
        & Simulation count & 64 & \multirow{2}{*}{-} & 32 \\
        & Gumbel sampling & 16 & & 16 \\
        \midrule
        \multirow{3}{*}{\textbf{Learning Setting}} & Learning rate & 0.02 to 0.005 & 0.1 & 0.02 \\
        & Batch size & 1,024 & 1,024 & 1,024 \\
        & Training steps & 100,000 & 150,000 & 100,000 \\
        \bottomrule
    \end{tabular}%
    }
    \caption{Training settings for 9x9 Go, 19x19 Go, and 19x19 Hex.}
    \label{tab:training_setting}
\end{table}

\subsection{9x9 Go}
\label{appendix:9x9Go_training}
In 9x9 Go training, we employed the Gumbel AlphaZero algorithm \cite{danihelka_policy_2022}, which can guarantee policy improvement even with a small number of simulations (e.g., 16, 32, 64) during training.
The training process is 100,000 training steps in total.
Initially, the learning rate is set to 0.02 for the first 70,000 steps.
Then, it is reduced to 0.01 for the subsequent 20,000 steps, from step 70,001 to 90,000.
For the final 10,000 steps, the learning rate is further decreased to 0.005.

To evaluate the performance of the models, we selected four KataGo models as baselines and conducted 500 games for each. The details of the selected KataGo models are listed in Table \ref{tab:kata_go_info}.
Each model was evaluated with a fixed thinking time of 2 seconds per move across a total of 2,000 games, ensuring fairness by having each model play 1,000 games as Black and 1,000 games as White.

\begin{table}[!h]
\centering
\resizebox{\columnwidth}{!}{%
\begin{small}
\begin{tabular}{lrr}
\toprule
 & \textbf{\# blocks} & \textbf{Elo ratings } \\
\midrule
\texttt{kata1-b6c96-s115648256} & 6 & 9644.0 $\pm$ 16.2 \\
\texttt{kata1-b10c128-s41138688} & 10 & 10144.5 $\pm$ 18.7 \\
\texttt{kata1-b10c128-s108710656} & 10 & 10683.7 $\pm$ 16.3 \\
\texttt{kata1-b15c192-s798345984} & 15 & 12033.9 $\pm$ 16.6 \\
\bottomrule
\end{tabular}%
\end{small}
}
\caption{The version of the selected KataGo models for 9x9 Go.}
\label{tab:kata_go_info}
\end{table}

\subsection{19x19 Go}
\label{appendix:19x19Go_training}
In 19x19 Go training, the models are supervised on a dataset containing 1,000,000 games played by 7-dan to 9-dan human Go players on Tygem~\cite{cho_tygemgo_2001}, a popular online Go platform for human players.

To evaluate the performance of the models, we compare them against four KataGo models used as baselines, conducting 500 games for each model. Details of the selected KataGo models are provided in Table \ref{tab:19x19_go_kata_go_info}.
For fairness, each model is evaluated under the same conditions, with a fixed thinking time of 5 seconds per move across a total of 2,000 games. Each model plays 1,000 games as Black and 1,000 games as White to ensure balanced comparisons.

\begin{table}[!h]
\centering
\resizebox{\columnwidth}{!}{
\begin{small}
\begin{tabular}{lrr}
\toprule
 & \textbf{\# blocks} & \textbf{Elo ratings } \\
\midrule
\texttt{kata1-b10c128-s41138688} & 10 & 10144.5 $\pm$ 18.7 \\
\texttt{kata1-b10c128-s108710656} & 10 & 10683.7 $\pm$ 16.3 \\
\texttt{kata1-b10c128-s46989824} & 10 & 10241.6 $\pm$ 18.4 \\
\texttt{kata1-b10c128-s56992512} & 10 & 10381.0 $\pm$ 17.5 \\
\bottomrule
\end{tabular}%
\end{small}
}
\caption{The version of the selected KataGo models for 19x19 Go.}
\label{tab:19x19_go_kata_go_info}
\end{table}

\subsection{19x19 Hex}
Two models, \texttt{6R} and \texttt{RRTRRT}, are trained for a total of 100,000 training steps in 19x19 Hex.
The training settings for 19x19 Hex are similar to 9x9 Go except following differences.
The learning rate is set to 0.02 and the simulation count is 32.

To evaluate performance, each model is tested with a fixed thinking time of 1 second per move against MoHex~\cite{arneson_monte_2010}, which serves as the baseline opponent.
A total of 1000 evaluation games are conducted for each model, with 500 games played as Black and 500 games as White to ensure fairness.

\section{Details and Additional Results for Global Information Ability}
\label{sec:appendix_global_info}
This section provides experimental details and additional results on the ability of \texttt{R3(RRT)} and \texttt{10R} to process global information, corresponding to Section 4.2, ``Global Information Ability of ResTNet," in the main text.
It includes experimental details on performance against the \textit{cyclic-adversary} \cite{wang_adversarial_2023}, along with results from board evaluation tasks and the details for recognition of ladder patterns.

\subsection{Cyclic-adversary}
\label{sec:appendix_cyclic_vit_adversary_atk}

\citep{wang_adversarial_2023} investigates the vulnerabilities of KataGo, state-of-the-art Go AI agents, by training a cyclic-adversary using victim-play, where the adversary learns by competing against a fixed version of the victim model.
The training follows a curriculum that progressively targets increasingly stronger versions of the victim model.
Their results show that cyclic-adversary can exploit specific weaknesses in KataGo to achieve high win rates ($>$ 77\% win-rate when KataGo uses 2048 search visits) despite being weaker than human amateurs in general gameplay.
By analyzing the attacks from adversarial policies, they found adversaries qualitatively win victims by creating circular patterns.

\subsection{Circular Patterns Game Collection}
\label{sec:games_containing_circular_patterns}

For the experiments described in the subsection ``Recognition of Circular Patterns" in the main text, we collected the openings from 24 games containing circular patterns from \citep{wang_adversarial_2023}\footnote{These games are available online: \url{https://goattack.far.ai/adversarial-policy-katago} ~\cite{wang_adversarial_2023} }.
The 24 openings are shown in Figure \ref{fig:24opening}.

\subsection{Board Evaluation in Circular Patterns}
\label{sec:appendix_circular_pattern_bv}

The performance of \texttt{R3(RRT)} and \texttt{10R} in recognizing circular patterns is evaluated using the mean squared error (MSE) between the ground truth and the network predictions, where a lower MSE indicates better performance.
Based on the pre-trained models for 19x19 Go, we add an extra \textit{board evaluation} head for predicting the ownership of each position on the board.
These models were then tested to determine their ability to recognize circular patterns by evaluating ownership among the 24 games ~\cite{wang_adversarial_2023}.
The evaluations were conducted at the endgame stage to ensure confirmed ownership of territories.

Table \ref{tab:circular_pattern_19x19_bvresult} lists the MSE of board evaluation results for 24 games with circular patterns provided by \cite{wang_adversarial_2023}.
Figures \ref{fig:id_1_6_cyc_bv}-\ref{fig:id_19_24_cyc_bv} show the evaluation results.
For each row of subfigures, the stones marked in red in subfigure (a) highlight the areas of interest.
The board evaluation likelihoods predicted by \texttt{10R} and \texttt{R3(RRT)} are displayed in subfigures (c) and (d), respectively, while the ground truth is depicted in subfigure (b).

\subsection{Defending Cyclic-adversary}
\label{sec:appendix_adversary_atk}
In the subsection ``Recognition of Circular Patterns" in the main text, \texttt{10R} and \texttt{R3(RRT)} play against the cyclic adversary.
In this experiment, \texttt{10R} and \texttt{R3(RRT)} are required to recognize these circular patterns and play correctly with a sequence of moves to avoid being attacked; otherwise, it will lose the game.
\texttt{10R} and \texttt{R3(RRT)} each play 30 games against the cyclic-adversary for each opening. With 24 openings, this results in a total of 30 × 24 games against the cyclic-adversary.
Table \ref{tab:circular_pattern_19x19_defendresult} shows the probabilities of being attacked by the cyclic-adversary for each opening across 24 games, where a lower attack rate indicates a more robust model in defending against the cyclic-adversary.

In summary, the results show that \texttt{R3(RRT)} demonstrates superior performance in handling circular patterns compared to \texttt{10R}.
Moreover, this suggests that the repeating patterns of \texttt{RRT} significantly enhance the model's ability to acquire both local and global knowledge and successfully defend against the cyclic-adversary.
Given that \texttt{R3(RRT)} model was trained solely on human game records, we believe that if the models were trained using AlphaZero with self-play games, it could potentially further reduce the probability of being attacked.

\subsection{Recognizing Ladder Patterns}
\label{sec:appendix_recognizing_ladder_pattern}
For the experiments on identifying ladder patterns mentioned in the subsection ``Recognition of Ladder Patterns" of the main text, we collected an additional dataset from Tygem \cite{cho_tygemgo_2001}.
In each game, only one training data is kept, specifically the ladder sequences at the beginning.
The training dataset consists of 1,665,000 unique ladder patterns, while the evaluation dataset contains a total of 166,500 ladder patterns.
For pre-trained \texttt{R3(RRT)} and \texttt{10R}, a \textit{ladder head} is integrated into the network outputs.
During training, the backbone parameters of \texttt{R3(RRT)} and \texttt{10R} are frozen, and only the ladder heads are updated.
During the evaluation, the models predict an escape success if the output of the ladder head is greater than 0.5, and escape failure if the output is less than $-0.5$.
Values within the range of $(-0.5, 0.5)$ are classified as unknown and considered incorrect predictions.

\newpage

\begin{table*} [h]
    \centering
    \begin{tabular}{l|cc}
        \toprule
        & \textbf{\texttt{10R}} & \textbf{\texttt{R3(RRT)}} \\
        \midrule
        ID-1 & 2.74531 & \textbf{1.93483}\\
        ID-2 & \textbf{1.09310} & 1.59510\\
        ID-3 & 2.79342 & \textbf{1.13911}\\
        ID-4 & 3.75009 & \textbf{0.75788}\\
        ID-5 & 3.86130 & \textbf{1.62457}\\
        ID-6 & 3.76487 & \textbf{3.18798}\\
        ID-7 & 3.03171 & \textbf{1.03335}\\
        ID-8 & 1.39696 & \textbf{1.20176}\\
        ID-9 & 2.04695 & \textbf{0.97399}\\
        ID-10 & 3.86357 & \textbf{2.29924}\\
        ID-11 & 3.81551 & \textbf{0.59214}\\
        ID-12 & 2.49706 & \textbf{0.05018}\\
        ID-13 & 3.79895 & \textbf{0.08454}\\
        ID-14 & 3.95434 & \textbf{2.35646}\\
        ID-15 & \textbf{0.00372} & 0.05155\\
        ID-16 & \textbf{0.00220} & 0.00746\\
        ID-17 & 1.09637 & \textbf{0.55442}\\
        ID-18 & 2.98301 & \textbf{1.54970}\\
        ID-19 & 3.62889 & \textbf{2.38178}\\
        ID-20 & 3.66682 & \textbf{0.26839}\\
        ID-21 & 3.79069 & \textbf{0.70545}\\
        ID-22 & 3.66638 & \textbf{0.48432}\\
        ID-23 & 0.76690 & \textbf{0.67713}\\
        ID-24 & \textbf{0.00110} & 0.11166\\
        \midrule
        average & 2.58413 $\pm$ 0.59 & \textbf{1.06763} $\pm$ \textbf{0.37}\\
        \bottomrule
    \end{tabular}%
    \caption{The MSE of board evaluation in 24 games from cyclic-adversary.}
    \label{tab:circular_pattern_19x19_bvresult}
\end{table*}
\begin{table*} [h]   
    \centering
    \begin{tabular}{l|ccc}
        \toprule
        & \textbf{\texttt{10R}} & \textbf{\texttt{R3(RRT)}} \\
        \midrule
        Opening-1 & 66.67\% & \textbf{30.00}\% \\
        Opening-2 & 20.00\% & \textbf{16.67}\% \\
        Opening-3 & 93.33\% & \textbf{43.33}\% \\
        Opening-4 & 66.67\% & \textbf{46.67}\% \\
        Opening-5 & 90.00\% & \textbf{13.33}\% \\
        Opening-6 & 86.67\% & \textbf{43.33}\% \\
        Opening-7 & 23.33\% & \textbf{6.67}\% \\
        Opening-8 & 23.33\% & \textbf{3.33}\% \\
        Opening-9 & 90.00\% & \textbf{50.00}\% \\
        Opening-10 & 56.67\% & \textbf{33.33}\% \\
        Opening-11 & 90.00\% & \textbf{3.33}\% \\
        Opening-12 & 66.67\% & \textbf{3.33}\% \\
        Opening-13 & 96.67\% & \textbf{6.67}\% \\
        Opening-14 & 93.33\% & \textbf{16.67}\% \\
        Opening-15 & 80.00\% & \textbf{16.67}\% \\
        Opening-16 & 66.67\% & \textbf{36.67}\% \\
        Opening-17 & 66.67\% & \textbf{46.67}\% \\
        Opening-18 & 56.67\% & \textbf{36.67}\% \\
        Opening-19 & 93.33\% & \textbf{40.00}\% \\
        Opening-20 & 46.67\% & \textbf{13.33}\% \\
        Opening-21 & 70.00\% & \textbf{0.00}\% \\
        Opening-22 & 93.33\% & \textbf{3.33}\% \\
        Opening-23 & 56.67\% & \textbf{43.33}\% \\
        Opening-24 & 93.33\% & \textbf{26.67}\% \\
        \midrule
        Average & 70.44\% $\pm$ 3.34\% & \textbf{23.91}\% $\pm$ \textbf{3.13}\% \\
        \bottomrule
    \end{tabular}%
    \caption{The probability of being attacked by the cyclic-adversary for \texttt{10R}, and \texttt{R3(RRT)} in 24 openings.} 
    \label{tab:circular_pattern_19x19_defendresult}
\end{table*}

\newpage
\begin{figure*}[h]
\centering
\foreach \n in {1, ..., 24} {
    \begin{subfigure}[b]{0.164\textwidth}
        \centering
        \includegraphics[width=\textwidth]{figures/Experiment/opening/\n.png}
        \caption{Opening-\n}
        \label{fig:\n}
    \end{subfigure}
    \ifnum\n=24
    \else
        \ifnum\n=4 \par\medskip\fi
        \ifnum\n=8 \par\medskip\fi
        \ifnum\n=12 \par\medskip\fi
        \ifnum\n=16 \par\medskip\fi
        \ifnum\n=20 \par\medskip\fi
    \fi
}
\caption{24 openings with circular patterns for cyclic-adversary.}
\label{fig:24opening}
\end{figure*}

\newpage

\begin{figure*}[h]
\centering
\foreach \n in {1, ..., 6} {
    \begin{subfigure}[b]{0.9\textwidth}
        \centering
        \includegraphics[width=0.17\textwidth]{figures/Appendix/cyc_24_opening_bv/mark_X/\n.pdf}
        \hspace{0.005\textwidth}
        \includegraphics[width=0.17\textwidth]{figures/Appendix/cyc_24_opening_bv/ground_truth/\n.pdf}
        \hspace{0.005\textwidth}
        \includegraphics[width=0.17\textwidth]{figures/Appendix/cyc_24_opening_bv/10R/\n.pdf}
        \hspace{0.005\textwidth}
        \includegraphics[width=0.17\textwidth]{figures/Appendix/cyc_24_opening_bv/10RT/\n.pdf}
        \caption{ID-\n}
        \end{subfigure}
        \par\medskip
}
\caption{Recognizing circular patterns through board evaluations in the cyclic-adversary game collection (game IDs 1 to 6). Each subfigure, from left to right, shows the game with a circular pattern, the ground truth, and the board evaluation results for \texttt{10R} and \texttt{R3(RRT)}, respectively.}
\label{fig:id_1_6_cyc_bv}
\end{figure*}

\begin{figure*}[h]
\centering
\foreach \n in {7, ..., 12} {
    \begin{subfigure}[b]{0.9\textwidth}
        \centering
        \includegraphics[width=0.17\textwidth]{figures/Appendix/cyc_24_opening_bv/mark_X/\n.pdf}
        \hspace{0.005\textwidth}
        \includegraphics[width=0.17\textwidth]{figures/Appendix/cyc_24_opening_bv/ground_truth/\n.pdf}
        \hspace{0.005\textwidth}
        \includegraphics[width=0.17\textwidth]{figures/Appendix/cyc_24_opening_bv/10R/\n.pdf}
        \hspace{0.005\textwidth}
        \includegraphics[width=0.17\textwidth]{figures/Appendix/cyc_24_opening_bv/10RT/\n.pdf}
        \caption{ID-\n}
        \end{subfigure}
        \par\medskip
}
\caption{Recognizing circular patterns through board evaluations in the cyclic-adversary game collection (game IDs 7 to 12). Each subfigure, from left to right, shows the game with a circular pattern, the ground truth, and the board evaluation results for \texttt{10R} and \texttt{R3(RRT)}, respectively.}
\label{fig:id_7_12_cyc_bv}
\end{figure*}

\begin{figure*}[h]
\centering
\foreach \n in {13, ..., 18} {
    \begin{subfigure}[b]{0.9\textwidth}
        \centering
        \includegraphics[width=0.17\textwidth]{figures/Appendix/cyc_24_opening_bv/mark_X/\n.pdf}
        \hspace{0.005\textwidth}
        \includegraphics[width=0.17\textwidth]{figures/Appendix/cyc_24_opening_bv/ground_truth/\n.pdf}
        \hspace{0.005\textwidth}
        \includegraphics[width=0.17\textwidth]{figures/Appendix/cyc_24_opening_bv/10R/\n.pdf}
        \hspace{0.005\textwidth}
        \includegraphics[width=0.17\textwidth]{figures/Appendix/cyc_24_opening_bv/10RT/\n.pdf}
        \caption{ID-\n}
        \end{subfigure}
        \par\medskip
}
\caption{Recognizing circular patterns through board evaluations in the cyclic-adversary game collection (game IDs 13 to 18). Each subfigure, from left to right, shows the game with a circular pattern, the ground truth, and the board evaluation results for \texttt{10R} and \texttt{R3(RRT)}, respectively.}
\label{fig:id_13_18_cyc_bv}
\end{figure*}

\begin{figure*}[h]
\centering
\foreach \n in {19, ..., 24} {
    \begin{subfigure}[b]{0.9\textwidth}
        \centering
        \includegraphics[width=0.17\textwidth]{figures/Appendix/cyc_24_opening_bv/mark_X/\n.pdf}
        \hspace{0.005\textwidth}
        \includegraphics[width=0.17\textwidth]{figures/Appendix/cyc_24_opening_bv/ground_truth/\n.pdf}
        \hspace{0.005\textwidth}
        \includegraphics[width=0.17\textwidth]{figures/Appendix/cyc_24_opening_bv/10R/\n.pdf}
        \hspace{0.005\textwidth}
        \includegraphics[width=0.17\textwidth]{figures/Appendix/cyc_24_opening_bv/10RT/\n.pdf}
        \caption{ID-\n}
        \end{subfigure}
        \par\medskip
}
\caption{Recognizing circular patterns through board evaluations in the cyclic-adversary game collection (game IDs 19 to 24). Each subfigure, from left to right, shows the game with a circular pattern, the ground truth, and the board evaluation results for \texttt{10R} and \texttt{R3(RRT)}, respectively.}
\label{fig:id_19_24_cyc_bv}
\end{figure*}

\end{document}